\documentclass[10pt,twocolumn]{article}

\usepackage[margin=0.8in]{geometry}
\usepackage{mathptmx} % Times-like font
\usepackage{microtype}

\usepackage{amsmath,amssymb}

\usepackage{graphicx}
\usepackage{booktabs}
\usepackage{multirow}
\usepackage{tabularx}
\usepackage{subcaption}
\usepackage{float}
\usepackage{placeins}
\usepackage{threeparttable}
\usepackage{array}
\usepackage{makecell}

\newcolumntype{P}[1]{>{\raggedright\arraybackslash}p{#1}}

\makeatletter
\renewcommand*{\@fnsymbol}[1]{%
  \ifcase#1
    \or *% 1: equal contribution
    \or \Letter% 2: corresponding author
    \or \ddagger% 3
    \or \mathsection% 4
    \or \mathparagraph% 5
    \or \|% 6
    \or **% 7
    \else\@ctrerr
  \fi}
\makeatother

\usepackage{xcolor}
\definecolor{linkblue}{rgb}{0.212,0.490,0.741}

\usepackage{url}
\usepackage{authblk}
\usepackage{fancyhdr}
\usepackage[numbers,sort&compress]{natbib}
\usepackage{hyperref}
\usepackage{marvosym}

\hypersetup{
    colorlinks=true,
    linkcolor=linkblue,
    citecolor=linkblue,
    urlcolor=linkblue,
    pdftitle={High-Resolution Water Sampling via a Solar-Powered Autonomous Surface Vehicle},
    pdfauthor={Misael Mamani et al.}
}

\title{\vspace{-1.5em}\textbf{High-Resolution Water Sampling via a \\Solar-Powered Autonomous Surface Vehicle}\vspace{-0.5em}}

\author[1]{Misael Mamani}
\author[1]{Mariel Fernandez}
\author[1]{Grace Luna\thanks{These authors contributed equally to this work.}}
\author[1]{Steffani Limachi$^\ast$}
\author[1]{Leonel Apaza$^\ast$}
\author[1]{\\Carolina Montes-Dávalos}
\author[1]{Marcelo Herrera}
\author[2]{Edwin Salcedo\thanks{Corresponding author: e.r.salcedoaliaga@qmul.ac.uk}}

\affil[1]{Department of Mechatronics Engineering, Universidad Católica Boliviana ``San Pablo'', La Paz, Bolivia}
\affil[2]{School of Electronic Engineering and Computer Science\\ Queen Mary University of London, UK}

\date{} % remove date

\begin{document}
\maketitle

% ---------- Abstract ----------
\begin{abstract}
Accurate water quality assessment requires spatially resolved sampling, yet most unmanned surface vehicles (USVs) can collect only a limited number of samples or rely on single-point sensors with poor representativeness. This work presents a solar-powered, fully autonomous USV featuring a novel syringe-based sampling architecture capable of acquiring 72 discrete, contamination\allowbreak -minimized water samples per mission. The vehicle incorporates a ROS 2 autonomy stack with GPS-RTK navigation, LiDAR and stereo-vision obstacle detection, Nav2-based mission planning, and long-range LoRa supervision, enabling dependable execution of sampling routes in unstructured environments. The platform integrates a behavior-tree autonomy architecture adapted from Nav2, enabling mission-level reasoning and perception-aware navigation. A modular 6×12 sampling system, controlled by distributed micro-ROS nodes, provides deterministic actuation, fault isolation, and rapid module replacement, achieving spatial coverage beyond previously reported USV-based samplers. Field trials in Achocalla Lagoon (La Paz, Bolivia) demonstrated 87\% waypoint accuracy, stable autonomous navigation, and accurate physicochemical measurements (temperature, pH, conductivity, total dissolved solids) comparable to manually collected references. These results demonstrate that the platform enables reliable high-resolution sampling and autonomous mission execution, providing a scalable solution for aquatic monitoring in remote environments.
\end{abstract}


% ============================================================
% OPTIONAL – Graphical Abstract
% ============================================================
\begin{figure}[ht]
    \centering
    \includegraphics[width=\linewidth]{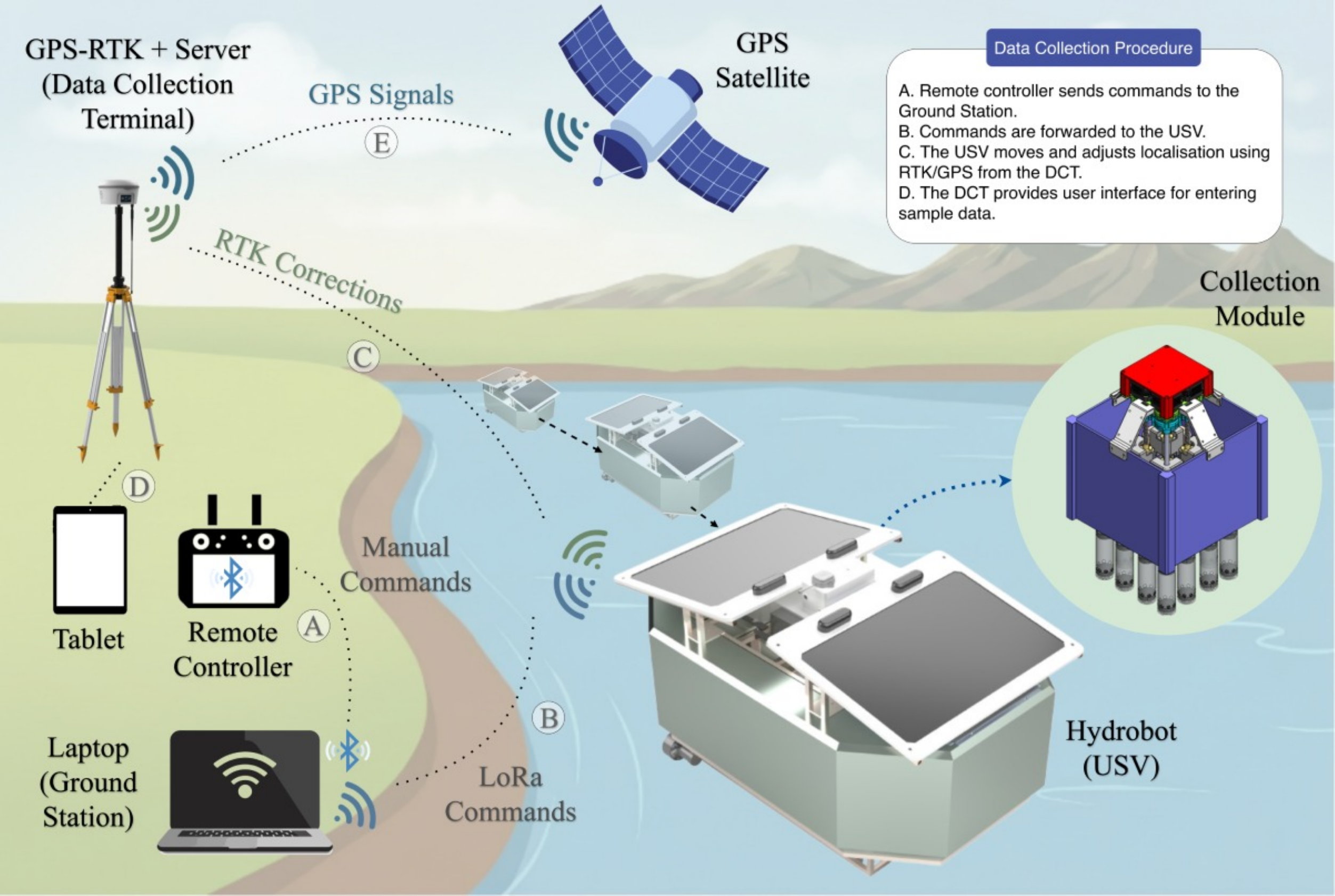}
    \caption{Overview of the proposed solar-powered autonomous USV and multi-syringe sampling system.}
\end{figure}

% ============================================================
% OPTIONAL – Highlights (arXiv-friendly)
% ============================================================
% \section*{Highlights}
% \begin{itemize}
%     \item Fully autonomous ROS~2-based USV integrating RTK-GPS, LiDAR, RGB-D perception, and mission-level behaviour trees.
%     \item Distributed micro-ROS embedded architecture coordinating 24 actuators across six modular sampling units.
%     \item Novel syringe-based design enabling 72 discrete, contamination-minimized samples per mission.
%     \item Solar-assisted multi-rail power system supporting extended deployments.
%     \item Field-proven in Achocalla Lagoon with 87\% waypoint tracking accuracy and reliable water-quality sampling.
% \end{itemize}

% \newpage
% \twocolumn

% ============================================================
% ===================== MAIN TEXT =============================
% ============================================================

\section{Introduction}
\label{sec:introduction}

Despite its essential role in sustaining life, safe drinking water remains out of reach for a significant proportion of the world’s population. Major contributors to water pollution include the discharge of untreated industrial and domestic waste, outdated mining practices, and the use of pesticides in agriculture \cite{unesco2018}. Water contamination, primarily from pathogenic microorganisms and chemical substances, causes diseases such as cholera, hepatitis A, typhoid fever, and acute diarrheal illnesses \cite{who2023}. Recent global estimates \cite{who2019} attribute 1.4 million preventable deaths and 74 million disability-adjusted life years (DALYs) to unsafe water, sanitation, and hygiene (WASH) services. In Latin America and the Caribbean, for example, the Pan American Health Organization (PAHO) reports that at least 28 million people in the region lack access to an improved water source, 83 million lack adequate sanitation facilities, and 15.6 million still practice open defecation, all of which increase the risk of waterborne diseases \cite{paho2017}.

To address persistent concerns over water quality, the international community adopted Sustainable Development Goal 6 (SDG 6), which requires countries to provide reliable evidence that drinking water supplies are safe, continuously available, and sourced from improved systems \cite{who2023}. PAHO/WHO guidelines operationalize these requirements by defining core microbiological and physicochemical indicators and by mandating rigorous sampling procedures to ensure analytical integrity \cite{paho2017guidelines}. Manual grab sampling, however, remains vulnerable to cross-contamination, offers limited temporal coverage, and often misses short-lived contamination events. Recent updates to international guidance highlight the need for higher-frequency, automated field sampling \cite{dbo2024}, underscoring a growing gap between recommended practices and what is operationally achievable with existing methods. This gap motivates the development of autonomous platforms capable of generating consistent, high-resolution water-quality data under real-world conditions.

With regard to water sampling automation, fixed-location samplers, such as the ISCO 3700C \cite{isco3700c_manual}, the ISCO 6712FR \cite{isco6712fr_manual}, and the WS-Series by YSI \cite{ysiWSseries_specsheet}, among others, stand out for their ability to collect water at specific depths and in precise quantities. However, they are constrained by their inability to adapt to changing environmental conditions or capture spatial variability across large water bodies. In contrast, the use of unmanned surface vehicles (USVs) represents a significant advancement over traditional methods. According to \cite{katsouras2024}, USVs enable more time-efficient sample collection, allow for real-time data acquisition, and provide access to remote or hazardous areas. One of their most relevant advantages is the reliability of the collected data, as the risk of cross-contamination is significantly reduced.
 
Despite technological advancements, several challenges persist in the practical deployment of USVs. Many of these systems still rely on basic autonomous navigation algorithms, which show notable limitations in performance and reliability. These shortcomings often cause frequent mission failures, forcing operators to revert to manual joystick control \cite{chang2021}. Moreover, most USVs cannot capture spatial and temporal variability effectively, as they typically collect only a few samples \cite{chang2021} or store all samples in a single container due to energy and load limitations \cite{ahmad2025}. Finally, data reliability may be compromised when sensors are embedded in USVs for continuous data acquisition \cite{katsouras2024}, as external disturbances, such as biological fouling, water currents, and waves, can interfere with the sampling process and variable measurements \cite{chen2025}, particularly when the vehicle remains submerged for prolonged periods. This highlights the ongoing need for offshore or laboratory-based analysis of samples collected by USVs to ensure higher measurement quality.

To address the limitations of existing sampling systems, we developed a solar-powered autonomous surface vehicle capable of automated water sampling from distributed locations across a water body. The platform features six modular sampling units, each containing 12 modified syringes (45 mL), enabling the collection of 72 discrete water samples per mission. These samples are transported to shore for in-situ measurement and registration of water quality parameters via a web-based data management system deployed on a bespoke control station. The USV is equipped with a comprehensive sensor suite—including GPS, IMU, LiDAR, and both visual and depth cameras, facilitating efficient navigation and obstacle avoidance. Onboard batteries and solar panels extend operational endurance, allowing long-duration deployments in remote aquatic environments and supporting the capture of spatial and temporal variability in water quality.

Therefore, the main contributions of this paper include:

\begin{itemize}
\item Design of a solar-powered autonomous USV with a modular mechanical architecture, supporting discrete, contamination-minimized water sampling in remote or unstructured environments. 

\item A scalable syringe-based sampling system capable of collecting 72 individual samples per mission, enhancing spatial resolution and enabling fine-grained environmental monitoring.

\item A ROS 2-based multilayer control architecture integrating GPS navigation, LiDAR-based obstacle avoidance, stereo vision, and long-range LoRa communication. 

\item A publicly accessible web platform \cite{hydrobot_web} for data management, allowing in-situ data registration and synchronization for missions conducted in offline environments.

\item Field validation in a Bolivian lagoon, demonstrating the system's autonomy, sampling accuracy, and energy efficiency through water quality measurements (e.g., pH, turbidity, conductivity) and verification of sampling volume precision. A video of the operation can be found at the following link \url{https://bit.ly/hydrobotbo}.
\end{itemize}

This paper is structured as follows. Section \ref{sec:literature-review} reviews methodologies for water sampling, covering both manual and automated systems, as well as recent developments in USVs for this purpose. Section \ref{sec:usv} describes the design and implementation of the USV, including its mechanical structure, electronic components, and navigation architecture. Section \ref{sec:water-sample-collection-mechanism} details the syringe-based water sampling mechanism. Section \ref{sec:experimental-results} presents the experimental results from field tests, highlighting the system’s performance in terms of autonomy, sampling reliability, and energy efficiency. Finally, Section \ref{sec:discussion} discusses the implications of the findings, and Section \ref{sec:conclusions} concludes the paper by summarizing the contributions and proposing future improvements.

% \FloatBarrier
\section{Literature Review}
\label{sec:literature-review}

This section reviews different methods of water sample collection, encompassing both manual procedures and automated systems. The latter include fixed platforms and USVs. Furthermore, we examine self-driving approaches that may provide a foundation for the development of the proposed vehicle.

\subsection{Manual water sampling}

Manual water sampling for routine inspections involves collecting water in a sterile container at a given time and location. While this approach may seem straightforward, it is limited in its ability to represent the spatial variability of large water bodies \cite{dbo2024}. It also becomes particularly challenging when sampling requires access to hazardous or contaminated areas. In such cases, composite samples may be obtained through scheduled procedures and with auxiliary tools such as peristaltic pumps \cite{agriculture2020}. Furthermore, under these circumstances, strict protocols—such as equipment sterilization, container pre-rinsing, the use of personal protective equipment, proper labeling, cold storage, and detailed documentation—are essential to ensure the integrity, traceability, and validity of collected samples \cite{dbo2024, usgs2006}.

In dynamic water bodies such as rivers and canals, professionals, including hydrologists, environmental engineers, and water quality technicians, often employ isokinetic sampling. This method ensures that water enters the sampling device at the same velocity and direction as the surrounding flow, thereby minimizing bias caused by flow speed. Similarly, equal increment sampling methods, such as EWI (Equal Width Increment) and EDI (Equal Discharge Increment), are used. In these methods, subsamples are collected along the cross-section of the water body at regular spatial or discharge intervals, respectively. These methodologies are widely recognized for their ability to accurately represent total discharge and flow-weighted average concentrations of pollutants or sediments \cite{usgs2006}.

\begin{table*}[t]
\caption{Overview of recent USV-based systems for water sampling and monitoring.}
\label{tab:usv_summary}
\centering
\scriptsize
\renewcommand{\arraystretch}{1.2}
\newcolumntype{Y}{>{\raggedright\arraybackslash}X}
\begin{tabularx}{\textwidth}{@{}P{1cm} P{1.5cm} P{2cm} P{1cm} P{2cm} P{1.5cm} P{2cm} P{1.5cm} P{1.5cm} @{}}
\toprule
\textbf{Author \& Year} & \textbf{Main Task} & \textbf{Measured Variables} & \textbf{Depth} & \textbf{Navigation (Sensors)} & \textbf{Main Processing Unit} & \textbf{Sampling Method} & \textbf{Sampling Volume} & \textbf{Vehicle Duration} \\
\midrule
Chen \cite{chen2025} 2025 & In-situ monitoring & pH, DO, EC, ORP & Surface (probe lowering; not specified) & Manual + waypoint-based (Pixhawk 6C, M9N GPS/GNSS, AIS, camera) & Pixhawk 6C + microcontroller & In-situ sensors only & Not required & Not specified \\
\hline
Lim \cite{lim2025} 2025 & Water sampling & Test kits (pH, TA, Hard, Nitrate, Nitrite, Fluoride, etc.) & Surface & Manual + waypoint-based (GPS, electronic compass, accelerometer) & ArduPilot APM 2.8 & Peristaltic pump + syringes & Not specified & Not specified \\
\hline
Ahmad \cite{ahmad2025} 2025 & Water sampling & pH, Temp, EC & 10–20 cm & Waypoint-based & ESP32 & 3 R385 pumps + 3D tank & Not specified & Not specified \\
\hline
Katsouras \cite{katsouras2024} 2024 & Water sampling / in-situ monitoring & pH, EC, Temp, DO, chlorophyll, Pb/Cu & Surface & Manual + waypoint-based (GO-SYS) & BlueBox + Raspberry Pi & 2 peristaltic pumps & 2000 ml (4 containers: 500 ml each) & 3 hours \\
\hline
Griffiths \cite{griffiths2022} 2022 & In-situ monitoring & Nitrate, EXO1 multiparametric, LI-190R & >30 cm & Not specified & Not specified & In-situ sensors only & Not required & 3 hours \\
\hline
Rashid \cite{rashid2022} 2022 & In-situ monitoring & EC, pH, depth & Not specified & Waypoint-based (GPS, compass, PID for disturbance) & Arduino Mega 2560 & In-situ sensors only & Not specified & Not specified \\
\hline
Chang \cite{chang2021} 2021 & Water sampling / water cleaning & pH, Pixy CMUcam5 & Surface & Manual + waypoint-based and obstacle avoidance (GPS, ultrasonic sensor) & 2× Arduino MEGA 2560 & R385 pump, electromagnetic valves & 60 ml (2 bottles: 30 ml each) & Not specified \\
\hline
Huang \cite{huang2021} 2021 & In-situ monitoring & YSI EXO1 (multi-param.) & 0–1000 cm & Waypoint-based (GPS, winchcam) & Arduino + ROS & Winch-controlled probe & Not required & Not specified \\
\hline
Bae \cite{bae2019} 2019 & Sediment sampling & Not specified (laboratory) & Sediment bed & Manual + waypoint-based (GPS, electronic compass)  & Arduino + Jetson Nano & Van Veen grab sampler & vol. max 1{,}630 cm$^3$ & Not specified \\
\hline
This work & Water sampling & pH, Temp, TDS, EC & 30 cm & Manual + autonomous (IMU, LiDAR, LoRa, RealSense Intel camera) & Jetson Nano + 2× ESP32 & 24 stepper motors (4 per module) + 3 syringes per motor & 72 discrete samples (45 ml) & Solar-powered + 2× LiFePO$_4$ battery (1 hour)  \\
\bottomrule
\end{tabularx}

\begin{tablenotes}
\item \textit{Abbreviations}: pH – acidity/alkalinity. EC – electrical conductivity. Temp – temperature. TA – total alkalinity. Hard – hardness. DO – dissolved oxygen. TDS – total dissolved solids. Pb – lead. Cu – copper.
\end{tablenotes}
\end{table*}

% \FloatBarrier
\subsection{Automated water sampling}

Common automated water sampling approaches include time-based sampling, in which samples are collected at fixed intervals; flow-proportional sampling, which adjusts collection according to water flow; and event-triggered sampling, which is activated by environmental changes such as turbidity or rainfall \cite{budai2020}. Among commercially available solutions, the Teledyne ISCO \cite{teledyneisco2025, isco3700c_manual, isco6712fr_manual} and YSI WS-Series \cite{ysiWSseries_specsheet} product lines are noteworthy. These systems enable programmable, unattended operation in both environmental and industrial settings. They employ non-contact peristaltic pumps connected to suction lines to extract water from surface bodies or pressurized conduits, and support both discrete and composite modes. Notably, Teledyne ISCO samplers incorporate refrigeration as well as automatic line purge and rinse functions before and after each cycle, which substantially reduce the risk of cross-contamination and improve sample preservation.

In addition, in-situ autoanalyzers—often equipped with autonomous or remotely controlled platforms and sensors (commonly referred to as the Internet of Things, IoT)—can continuously monitor parameters such as pH, conductivity, and nitrate concentration. Recent developments in IoT-based water quality monitoring integrate wireless sensor networks and machine learning algorithms to enable real-time analytics, anomaly detection, and predictive modeling of water parameters \cite{singh2024}. Although these advances have facilitated remote deployment, IoT-based systems remain limited in spatial representativeness, since a single location cannot capture the variability among different points within the same water body, and costs can increase substantially when these systems are installed in a distributed manner.

\subsection{Unmanned surface vehicles for water sampling}

USVs offer an innovative and dynamic solution that reduces human intervention while expanding spatial coverage in water monitoring activities. As summarized in Table~\ref{tab:usv_summary}, the most common configurations for water sampling with USVs employ peristaltic pumps \cite{lim2025, ahmad2025, katsouras2024, chang2021}. Spatial representativeness depends directly on the design of the sampling storage system. For example, Lim et al.~\cite{lim2025} equipped their USV with six syringes to collect samples across Ayer Keroh Lake, Malaysia; Katsouras et al.~\cite{katsouras2024} implemented four 500\,ml containers to obtain distributed samples in Koumoundourou Lake and the rivers Acheloos, Asopos, and Kifissos; and Chang et al.~\cite{chang2021} used two 30\,ml syringes for discrete storage. In the latter study, a pH sensor was integrated into the vehicle, operating continuously and activating the sampling pump only when the measured values exceeded a predefined range.

Beyond discrete sampling, recent research has explored systems in which water is not collected but rather measured directly through embedded sensors. For example, Griffiths et al.~\cite{griffiths2022} and Huang et al.~\cite{huang2021} implemented multiparameter probes capable of recording water quality parameters in real time at depths ranging from 30\,cm to 1\,m. Similarly, Katsouras et al.~\cite{katsouras2024} integrated continuous monitoring sensors on the underside of their vehicle, enabling acquisition of parameters such as pH, electrical conductivity, temperature, dissolved oxygen, chlorophyll, and metals (e.g., Pb and Cu) during navigation without interruption. Ahmad et al.~\cite{ahmad2025} designed a system with three peristaltic pumps: two drew water at 10\,cm and 20\,cm below the surface and directed it to a 3D-printed container housing sensors for temperature, pH, and electrical conductivity, while the third pump expelled the water to allow new measurements. As in the present project, we tested the integration of pH and temperature sensors but found that the measurements varied considerably, therefore, we excluded this approach.

Furthermore, some studies have extended the use of USVs to sediment collection, which is essential for identifying contaminants such as heavy metals, nutrients, and persistent organic compounds that accumulate at the water–sediment interface. Bae et al. \cite{bae2019} presented a USV equipped with a Van Veen sampler capable of collecting a single sediment sample with a maximum volume of 1630 cm\textsuperscript{3}. However, future research should address the current lack of spatial representativeness, as existing approaches remain restricted to single sampling points. Indeed, manual sediment collection presents considerable technical challenges due to depth, operator stability, and bottom conditions, highlighting an important research gap. 

\subsection{Autonomous navigation strategies in unmanned surface vehicles}

The levels of autonomy in USVs are primarily defined by the degree of human intervention required during their operation. At the most basic level, remote control, the vehicle depends on a pilot who manually guides it within the pilot’s visual range. As shown in Table \ref{tab:usv_summary}, most of the USVs analyzed include manual operation as an alternative navigation mode \cite{lim2025, katsouras2024, chang2021, bae2019}, often used as a fallback if the primary control system fails. A higher level corresponds to teleoperated systems, in which control remains human driven but is carried out through onboard sensors and cameras that allow the operator to monitor the vehicle’s status and perspective, even beyond the line of sight \cite{elmokadem2021}.

Semi-autonomy introduces a degree of independence, in which the operator defines general objectives such as routes or waypoints (also referred to as waypoint navigation), while the vehicle autonomously moves toward them with enhanced decision-making capabilities for collision avoidance and path planning \cite{elmokadem2021}. Examples of this approach are reported in \cite{lim2025, ahmad2025, bae2019}, where waypoint navigation is implemented using accelerometers and GPS/IMU. In a complementary study, \cite{chang2021} describes a semi-autonomous system that, in addition to waypoint navigation, incorporates ultrasonic sensors, enabling the vehicle to advance until an obstacle is detected and then adjust its trajectory. Similarly, \cite{rashid2022} proposes waypoint navigation enhanced with a proportional–integral–derivative (PID) controller on the motors to compensate for external disturbances. More recently, \cite{chen2025} developed a USV for in situ water quality monitoring that integrates GPS-based autonomous navigation, long-range communication, and an AIS module, demonstrating accurate field performance but still requiring improvements in obstacle avoidance and energy autonomy through the addition of sensors and solar panels.

At the most advanced level is full autonomy, where the vehicle receives a high-level task or mission definition from the user and executes it independently. In this case, the USV makes real-time decisions based on sensor data and adapts to changing environmental conditions and obstacles. For instance, researchers in \cite{salcedo2024} propose a USV equipped with a deep learning–based steering angle estimation (SAE) system for duckweed collection, navigation, and obstacle avoidance in infested water bodies. In the field of water sampling and monitoring, this level of autonomy remains an open research direction to which the present work aims to contribute.

\begin{figure*}[ht]
    \centering
    
    \begin{subfigure}[b]{0.48\linewidth}
        \centering
        \includegraphics[width=\linewidth]{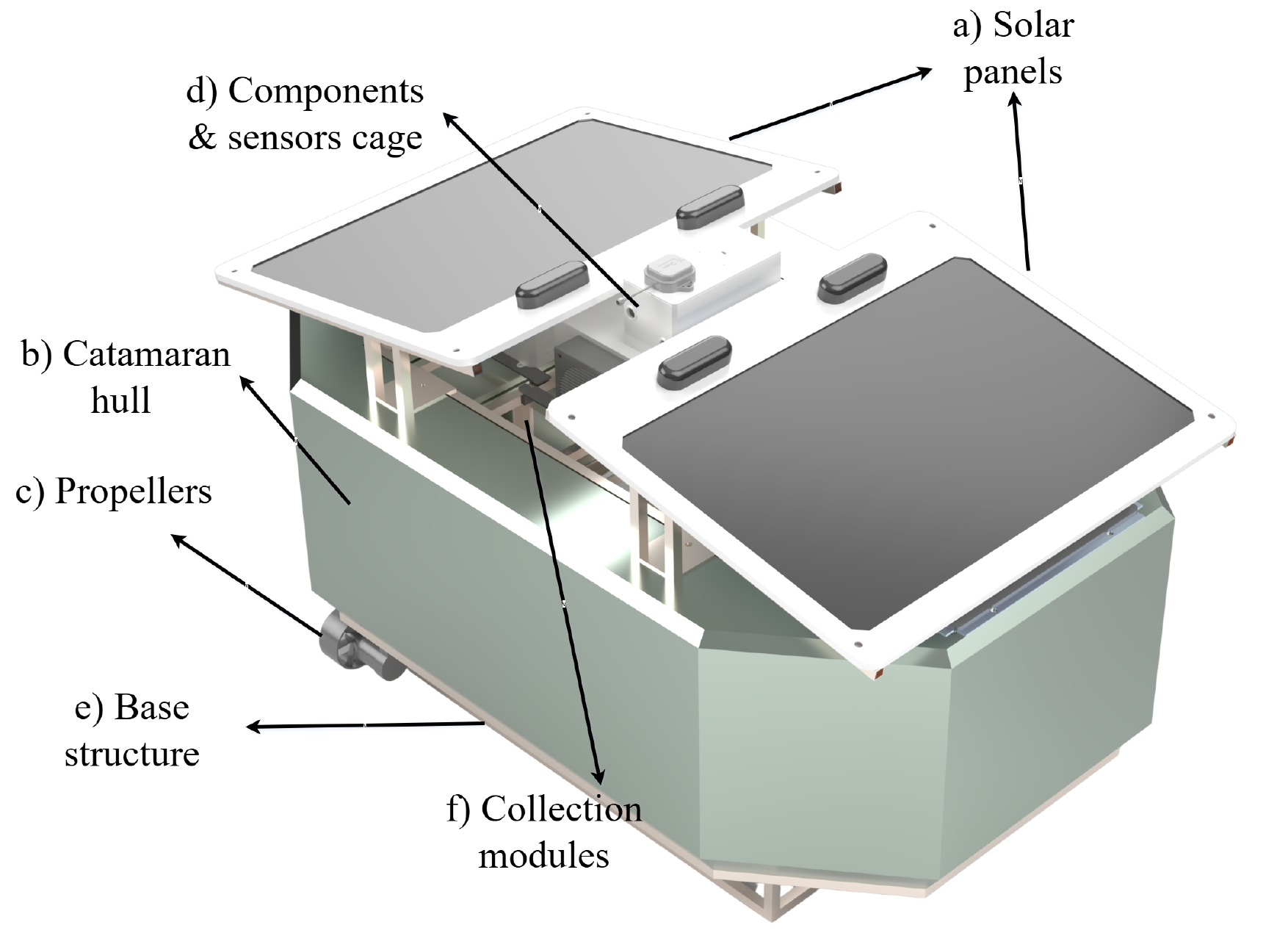}
        \caption{} 
        \label{fig:usv-cad}
    \end{subfigure}
    \hfill
    \begin{subfigure}[b]{0.48\linewidth}
        \centering
        \includegraphics[width=\linewidth]{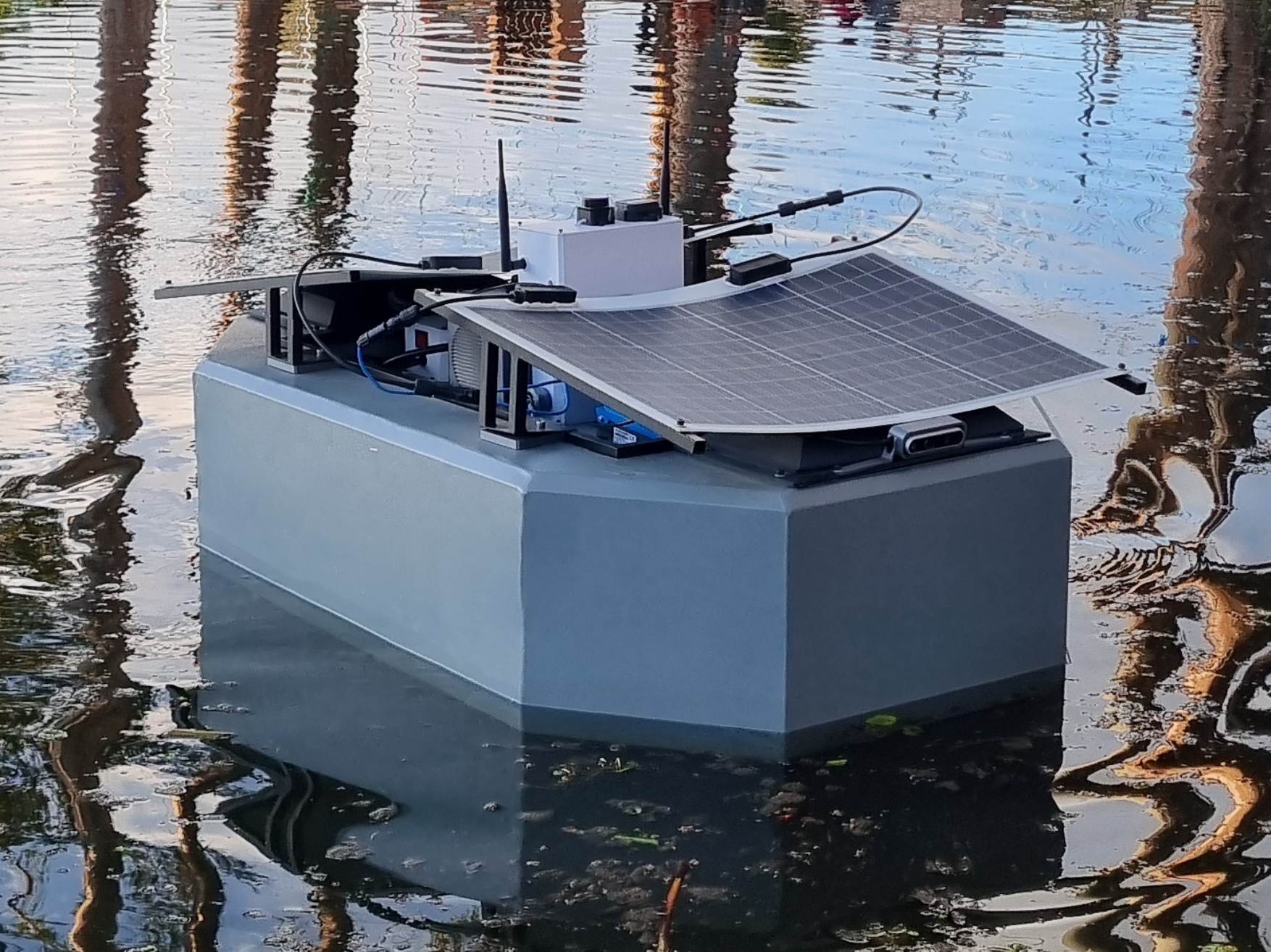}
        \caption{} 
        \label{fig:usv-prototype}
    \end{subfigure}

    \vspace{1em}

    \begin{subfigure}[b]{0.48\linewidth}
        \centering
        \includegraphics[width=\linewidth]{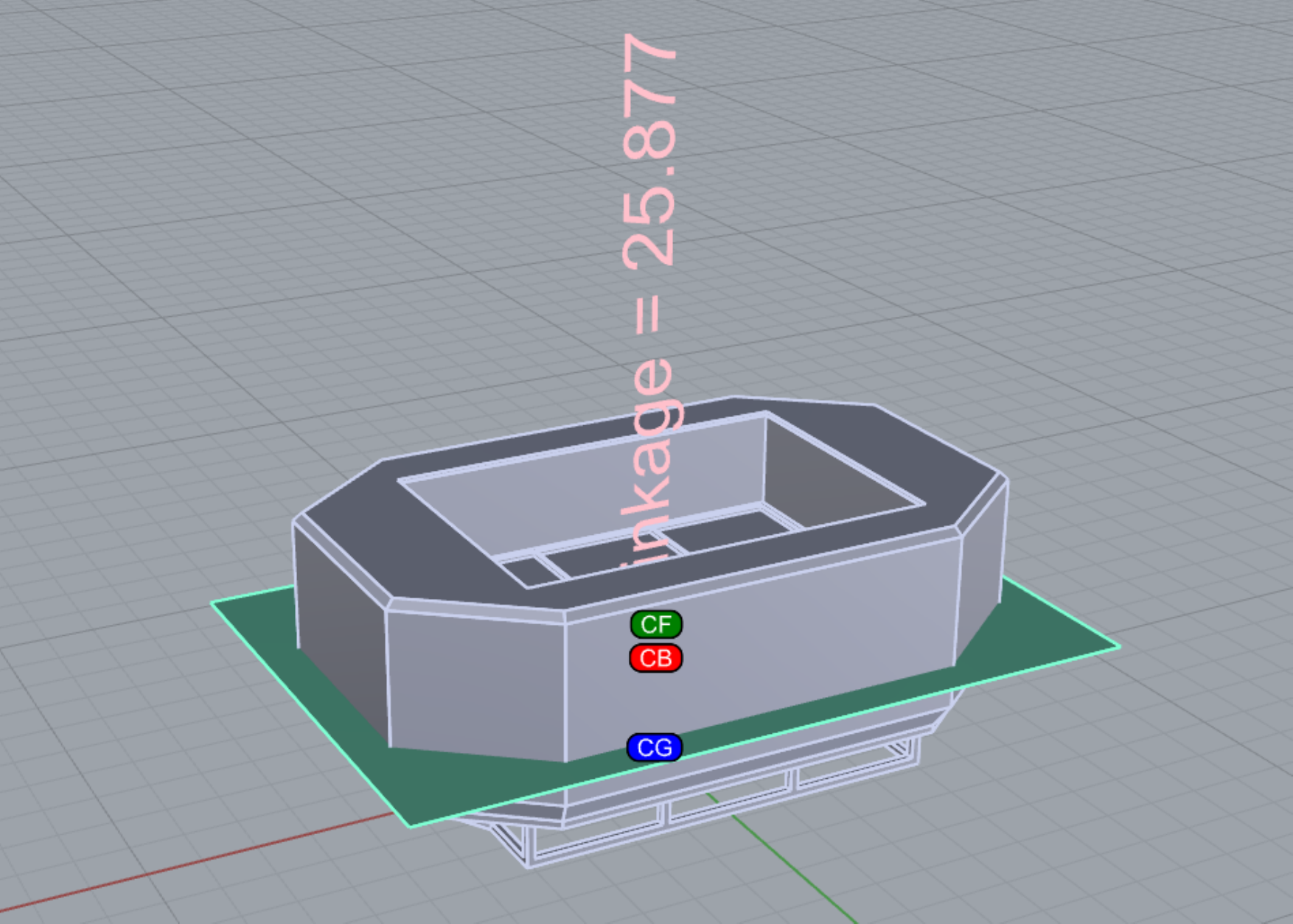}
        \caption{}
        \label{fig:rhino-analysis}
    \end{subfigure}
    \hfill
    \begin{subfigure}[b]{0.48\linewidth}
        \centering
        \includegraphics[width=\linewidth]{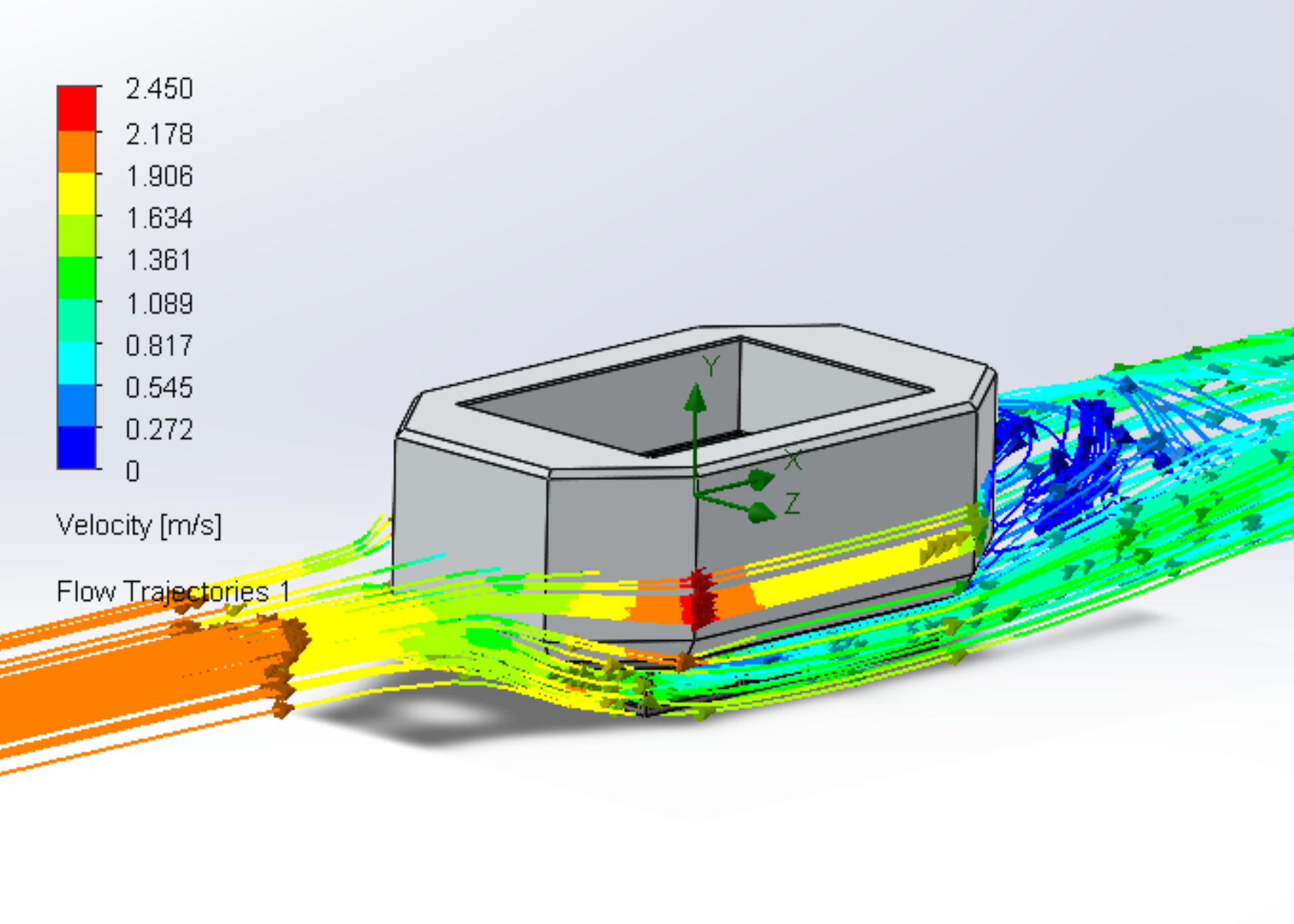}
        \caption{} 
        \label{fig:solidworks-analysis}
    \end{subfigure}

    \caption{
        (A) CAD rendering of the USV for water sampling. 
        (B) Physical prototype of the USV operating in Achocalla, La Paz, Bolivia. 
        (C) CFD analysis using Rhino and Orca3D, showing an immersion depth of approximately 25\,cm measured from the sampling modules. 
        (D) Flow analysis using SolidWorks Simulation Tools representing hydrodynamic drag as a function of water velocity.
    }
    \label{fig:combined-usv}
\end{figure*}

\begin{table*}[t]
\centering
\scriptsize

\caption{List of hardware components used in the unmanned surface vehicle and remote control.}
\label{tab:components}

\begin{tabularx}{\textwidth}{@{}P{0.4cm} P{2.5cm} P{1.5cm} P{6.5cm} P{1.8cm} P{2cm}@{}} % l|p{3.2cm}|p{3cm}|p{6.5cm}|p{2.5cm}|p{3cm}
\toprule
 & \textbf{Component} & \textbf{Purpose} & \textbf{Characteristics} & \textbf{Size (mm)} & \textbf{Energy consumption} \\
\midrule

% --- USV ---
\multirow{13}{*}{\rotatebox{90}{\textbf{USV}}} 
& Jetson Orin Nano & Processing & 6-core ARM Cortex-A78AE, 8 GB LPDDR5, 1024-core GPU + 32 Tensor Cores & 100\,×\,80\,×\,29 & 9--20 V DC, 1.7 A (15 W max) \\
& ESP32 DevKit & Control & Dual-core Xtensa LX6 @ 240 MHz, Wi-Fi + BLE & 58\,×\,28\,×\,12 & 3.3 V, 0.5 A \\
& Intel RealSense D435 & Sensing & RGB-D camera, USB 3.1, depth range \textasciitilde10 m & 90\,×\,25\,×\,25 & 5 V, 0.35 A \\
& SimpleRTK2B Starter Kit LR & Localisation & RTK GNSS with u-blox ZED-F9P, 2 LR radios, 2 GNSS antennas & 56\,×\,40\,×\,20 & 3.3--5.5 V, 0.2 A/module \\
& LoRa SerialKit & Communication & UART, 915 MHz, 10 km LOS range, TX 20 dBm & 40\,×\,30\,×\,5 & 3.3 V, 120 mA (TX), 15 mA (RX) \\
& VectorNav VN-100 Rugged & IMU / AHRS & 9-DOF IMU, 32-bit processor, temperature-compensated & 24\,×\,22\,×\,3 & 3.2--5.5 V, 40 mA \\
& LiDAR & Mapping & 360\textdegree{} scan, 0.12--12 m, 4500 pts/s & 72\,×\,72\,×\,50 & 5 V, 260 mA \\
& LiFePO$_4$ batteries & Power & 25.6 V, 50 A, >2700 cycles, built-in BMS & 330\,×\,170\,×\,215 & N/A \\
& Voltage regulator 1 & Stabilisation & DC-DC step-down, 4.5--40 V input, 1.23--37 V output & 43\,×\,21\,×\,13 & N/A \\
& Voltage regulator 2 & Stabilisation & Buck converter, 4--40 V in, 1.25--36 V out, 4.8 A & 45\,×\,21\,×\,14 & N/A \\
& Solar charge controller & Stabilisation & SmartSolar MPPT 150/100-Tr, 150 V in, 100 A out & 216\,×\,295\,×\,103 & <1 W (self) \\
& Solar panels & Power & 50 W, 18.9 V, monocrystalline, semi-flexible & 580\,×\,540\,×\,25 & N/A \\
& Brushless motors & Movement & 24 V, 900 W, 22 lbf thrust, brushless & 99\,×\,70\,×\,50 & 24 V, 37.5 A (full load) \\
% If you want ESCs here, uncomment the next line and increase the 13 above to 14
& ESC brushless controllers & Motor control & 24 V, 100 A, bidirectional, regenerative braking & 85\,×\,60\,×\,35 & N/A \\

\midrule

% --- Sampling module ---
\multirow{7}{*}{\rotatebox{90}{\shortstack{\textbf{Sampling}\\\textbf{module}}}}
& ESP32 DevKit & Control & Dual-core Xtensa LX6 @ 240 MHz, Wi-Fi + BLE & 58\,×\,28\,×\,12 & 3.3 V, 0.5 A \\
& SX1509 I/O expander & GPIO expansion & 16 GPIOs, I\textsuperscript{2}C interface, keypad/LED engine, 1.2--3.6 V logic & 20\,×\,20\,×\,2 & 3.3 V, 1 mA \\
& Motor NEMA17 & Movement & 1.8\textdegree{}/step, 1.65 N\,m, 1.7 A, bipolar, 2-phase & 42\,×\,42\,×\,40 & 12--24 V, 1.7 A \\
& A4988 driver & Motor control & Stepper driver, microstepping 1/16, 8--35 V, 2 A/coil & 20\,×\,15\,×\,10 & 3.3--5 V, 2 A \\
& Limit switch & Detection & Mechanical, normally open/closed & 12.8\,×\,5.8\,×\,6.2 & N/A \\
& Fuse & Protection & Glass cartridge, up to 250 V, 0.1--10 A & 30\,×\,6 & N/A \\
& Medical syringe & Sampling & 50 mL, plastic body, smooth plunger, hermetic seal & 150\,×\,30\,×\,30 & N/A \\

\midrule

% --- Remote control ---
\multirow{5}{*}{\rotatebox{90}{\shortstack{\textbf{Remote}\\\textbf{control}}}}
& ESP32 DevKit & Control + comm. & Dual-core Xtensa LX6 @ 240 MHz, Wi-Fi + BLE & 58\,×\,28\,×\,12 & 3.3 V, 0.5 A \\
& Joystick module & Movement & XY biaxial joystick, 2\,×\,10 k$\Omega$ potentiometers & 32\,×\,26\,×\,20 & 5 V \\
& Lithium batteries & Power & 3.7 V, 4800 mA & 18\,×\,65 & N/A \\
& OLED display 1.3'' & Visualisation & 128×64 px, SSD1306, I\textsuperscript{2}C/SPI, high contrast & 35\,×\,35\,×\,5 & 3.3--5 V, 20--30 mA \\
& Voltage regulator & Stabilisation & Step-down, 2--37 V, 3 A & 43\,×\,20\,×\,14 & N/A \\

\midrule

% --- DCT ---
\multirow{3}{*}{\rotatebox{90}{\textbf{DCT}}} 
& Raspberry Pi 4 Model B & Server & Quad-core ARM Cortex-A72 @ 1.5 GHz, 4 GB RAM & 85\,×\,56\,×\,17 & 5 V, 3 A \\
& SimpleRTK2B Starter Kit LR & Localisation & RTK GNSS with u-blox ZED-F9P, 2 LR radios, 2 GNSS antennas & 56\,×\,40\,×\,20 & 3.3--5.5 V, 0.2 A/module \\
& Power bank 20{,}000 mAh & Power & LiPo cells, USB-C PD + USB-A (22.5 W) output & 150\,×\,72\,×\,27 & N/A \\

\bottomrule
\end{tabularx}
\end{table*}

\begin{figure*}
\centering
\includegraphics[width=\linewidth]{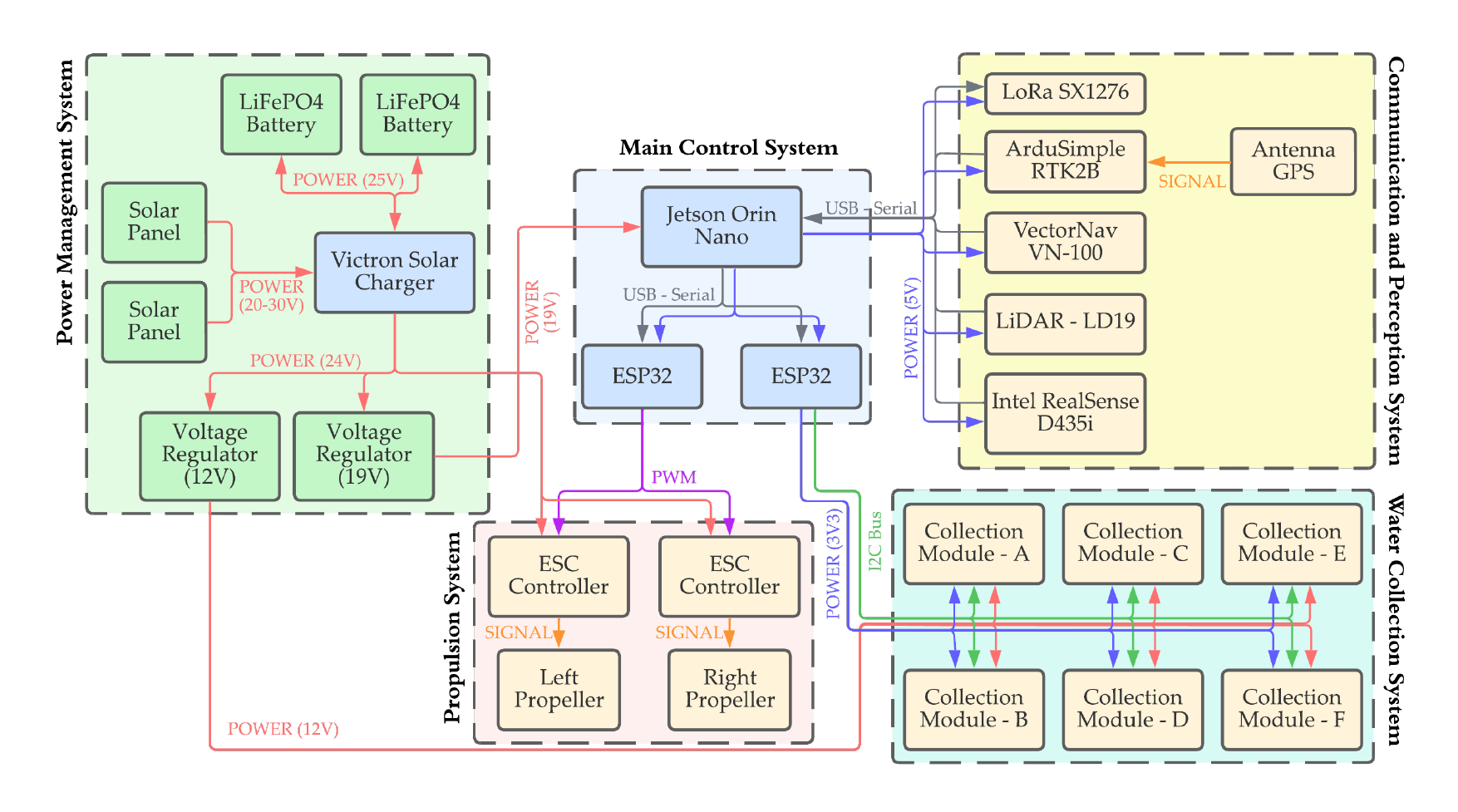}
\caption{Block diagram of the USV's electronic system, showing interconnections, power lines, and communication buses, and highlighting the integration of control, perception, communication, and power management subsystems.}
\label{fig:electronic_diagram}
\end{figure*}

\section{Design and Implementation of the Unmanned Surface Vehicle}
\label{sec:usv}

The present section describes the design and implementation process of the autonomous, solar-powered unmanned surface vehicle (USV), encompassing its mechanical, electronic, and software subsystems. The development followed an integrated workflow aimed at achieving robust performance in autonomous water sampling operations. A distributed control architecture was implemented to coordinate sensors, actuators, and communication modules, while the autonomy framework, built on ROS 2 Humble and micro-ROS, enabled the vehicle to perform fully autonomous sampling missions.

\subsection{Mechanical design}

The mechanical development used CAD tools and simulation environments to refine the hull geometry and internal component layout. As shown in Figure~\ref{fig:usv-cad}, a comprehensive 3D model was created in \textsc{SolidWorks}, serving as the backbone for iterative improvements informed by hydrostatic and CFD (Computational Fluid Dynamics) analyses conducted in \textsc{Rhino} with the \textsc{Orca3D} plugin. These design iterations were instrumental in meeting two critical requirements: a maximum static draft of 35\,cm to ensure navigation in shallow waters and stable operation under river flow velocities of up to 1.5\,m\,s$^{-1}$, given that the USV must carry the components listed under the USV section in Table \ref{tab:components}. The final configuration adopted a modified catamaran layout, balancing hydrodynamic drag, deck area, and overall stability, as shown in Figure~\ref{fig:solidworks-analysis}.

The hull features a 1.20 m $\times$ 0.30 m pontoon structure fabricated from 6061 aluminum, bolted to a stainless-steel cage constructed from 15 mm $\times$ 15 mm square tubing. This frame forms a ``floating cube'' architecture that enhances structural integrity and facilitates integration of batteries and propulsion elements. Two rectangular recesses are machined into the hull to house the LiFePO$_4$ battery packs. Positioning the twin thrusters low and centrally within the frame enhances yaw authority and reduces pitching moments during rapid maneuvers or when encountering waves. Pressure-diffusion simulations verified that the hull design remains watertight and structurally sound during sharp turns and under moderate wave impacts, thereby fulfilling the immersion and safety targets established at the outset of the project.

Within the central frame, a 1.00\,m $\times$ 0.50\,m modular bay houses six water-collection modules arranged in a 2 $\times$ 3 matrix. The attachment points for these modules are spaced at 0.40\,m intervals, a design choice that reduces lateral flex and damps vibrational loads when all mechanisms are actuated simultaneously. Above the frame, a composite tray measuring 1.20\,m $\times$ 0.60\,m supports the onboard computer, power electronics, and two solar panels, each measuring 0.50\,m $\times$ 0.30\,m. Dedicated sensor mounts, many of which are 3D-printed, maintain unobstructed fields of view for the LiDAR and stereo/depth cameras, while a slide-in rail system facilitates sensor-module replacement within minutes using standard tools. This modularity enables rapid reconfiguration across mission profiles while reducing maintenance downtime.

All stainless-steel joints were assembled using TIG (Tungsten Inert Gas) welding to ensure clean, pressure-tight seams, while the aluminum pontoons were sealed via precision laser welding; each unit underwent individual pressure testing prior to final assembly. A 150\,$\mu$m powder coat, cured at 180\,$^\circ$C, was applied to all metallic surfaces, providing corrosion and abrasion resistance suitable for sediment-rich riverine environments. The assembly process was coordinated, with concurrent machining, welding, and coating workflows reducing the total build time to four weeks. The result is a fully integrated hull structure weighing 105\,kg (including all electronics), ready for field deployment.

\subsection{Electronic design}

The electronic architecture of the USV is organized into four major subsystems: power management, central control and processing, communication and perception, and the water-collection system, as illustrated in Figure \ref{fig:electronic_diagram}. These subsystems are interconnected via standardized protocols and carefully regulated power lines to ensure reliable operation under challenging field conditions.

At the core of the control system is an NVIDIA Jetson Orin Nano single-board computer, which serves as the main processing hub for data fusion, navigation, and sensor integration. The Jetson is powered by a dedicated 19\,V DC rail regulated by a high-efficiency DC-DC converter. Two ESP32 microcontrollers are connected to the Jetson via USB-to-serial links, acting as real-time bridges for low-level actuator control and sensor interfacing. Communication between the ESP32 units and the sampling modules occurs over an I\textsuperscript{2}C bus, operating at 3.3\,V logic levels, which supports distributed control and monitoring across the six water-collection modules.

The perception and communication subsystem comprises a suite of high-precision sensors and telemetry devices, each powered by a regulated 5\,V supply. The ArduSimple RTK2B GNSS module provides centimeter-level positioning accuracy and interfaces with an external antenna for enhanced satellite reception. Attitude and orientation data are supplied by the VectorNav VN-100 IMU, while environmental awareness is provided by an Intel RealSense D435i RGB-D camera and an LD-19 360$^\circ$ LiDAR, capable of detecting obstacles up to 12\,m from the vessel. All sensors interface with the Jetson via USB or serial protocols, supporting high-throughput, low-latency data acquisition for autonomous navigation.

For long-range telemetry and remote command, the system integrates a LoRa SX1276 transceiver operating at 915\,MHz, enabling robust communication beyond visual line of sight. Data and control signals from the perception and collection modules are routed and isolated to prevent interference, with communication lines employing differential signaling or shielded cabling where appropriate. Propulsion is managed by two electronic speed controllers (ESCs), each driving a thruster. The ESCs receive PWM control signals from the ESP32 microcontrollers and are powered by a 12\,V rail derived from a 24\,V supply regulated by a high-current DC-DC converter. All system components share a star-ground topology, and critical circuits are protected by blade fuses to enhance electrical safety and simplify field maintenance.

The power management subsystem comprises two parallel-connected 25.6\,V LiFePO$_4$ batteries, charged in the field via dual 50\,W monocrystalline solar panels and a Victron MPPT solar charge controller. Power is distributed across four voltage levels: 24\,V for the core supply, 19\,V for the Jetson, 12\,V for propulsion, and 5\, and 3.3\,V for sensor and logic circuitry. This hierarchy of voltage rails, combined with the use of high-efficiency converters, maximizes system uptime and ensures consistent operation under varying environmental conditions.

\subsection{Autonomy framework}
\label{sec:software-integration}

The USV's autonomy and perception system is built on the ROS 2 Humble middleware, which provides a modular framework for integrating sensors, actuators, and communication modules. Each component interfaces through official ROS 2 packages or custom nodes, all interconnected via a message-driven architecture managed by the Jetson Orin Nano. USB–serial connections link the ESP32 microcontrollers to the Jetson, while micro-ROS agents integrate the embedded nodes into the ROS 2 computational graph. Status and diagnostic messages are published locally for onboard monitoring and, when necessary, transmitted via the LoRa channel to the ground station for remote supervision. In parallel, all mission-related sensor data and collected water sample information are transmitted to and stored on a remote server (detailed in Section \ref{sec:dct}), enabling structured post-mission analysis and dataset generation for environmental research. This hierarchical software architecture ensures reliable, maintainable, and extensible operation suitable for advanced autonomous surface-vehicle tasks. In what follows, we describe the perception pipeline, communication system, and self-driving approach that constitute the core of the autonomy framework.

\subsubsection{Perception pipeline} \label{subsec:perception-pipeline}

\begin{figure}
\centering
\includegraphics[width=\linewidth]{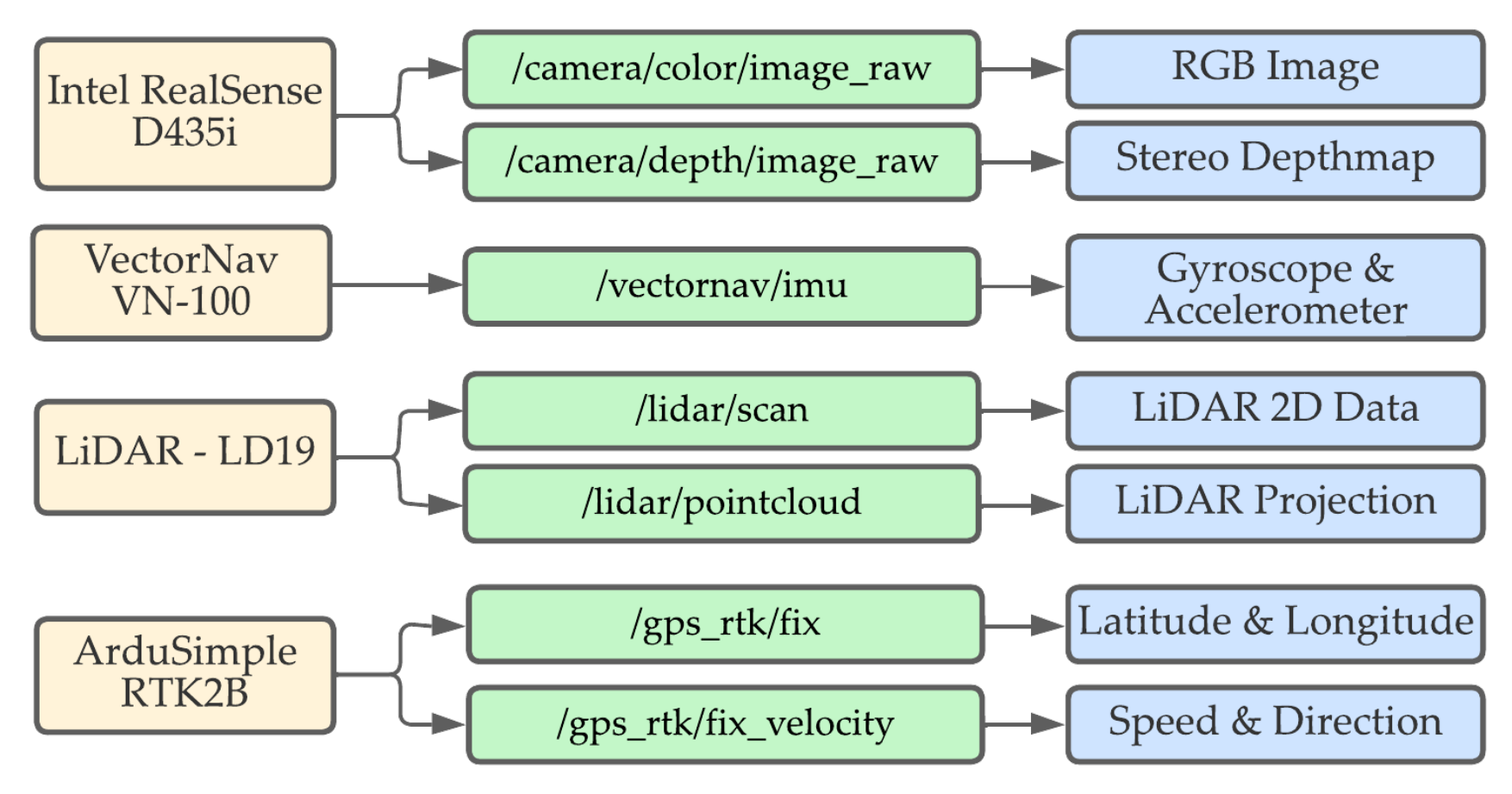}
\caption{ROS 2 topics and data published by perception sensors.}
\label{fig:perception_pipeline}
\end{figure}

Environmental perception uses official ROS 2 nodes to integrate onboard sensors, as illustrated in Figure \ref{fig:perception_pipeline}. This perception layer fuses visual, inertial, LiDAR, and GNSS data to build a coherent spatial representation of the environment. The Intel RealSense D435i outputs synchronized RGB and depth information for dense 3D reconstruction and obstacle detection. The VectorNav VN-100 streams high-frequency inertial and orientation data, while the ArduSimple RTK2B receiver provides precise global positioning and velocity measurements through RTK corrections. Complementing these, the LD-19 LiDAR generates two-dimensional scans and point-cloud projections for accurate mapping and collision avoidance.

Once acquired, all sensor data are temporally synchronized and spatially aligned using the ROS 2 transform (TF) framework, ensuring consistent reference frames across the system. These data streams are fused by higher-level navigation and mapping nodes to enable localization, path planning, and situational awareness, while the modular perception design ensures robustness and adaptability in dynamic aquatic environments. This setup provides the sensory foundation required for reliable autonomous operation.

\subsubsection{Mission planning and autonomy}

The mission planning framework was implemented using the ROS 2 Navigation Stack (Nav2), originally designed for ground robots operating in 2D environments and adapted here for the USV. This setup integrates global path planning, local control, and recovery behaviors within a behavior tree–based architecture, enabling the USV to reach predefined sampling points while avoiding obstacles. Simulations were conducted using TurtleBot3 packages in Gazebo, where the simulator published the robot’s pose in the $\text{base\_link}$ frame, along with gyroscope and simulated encoder data (although the physical implementation does not rely on wheel encoders, as detailed in Section \ref{sec:odometry}).

\begin{figure}[H]
    \centering
    \begin{subfigure}[b]{0.5\linewidth}
        \centering
        \includegraphics[width=\linewidth]{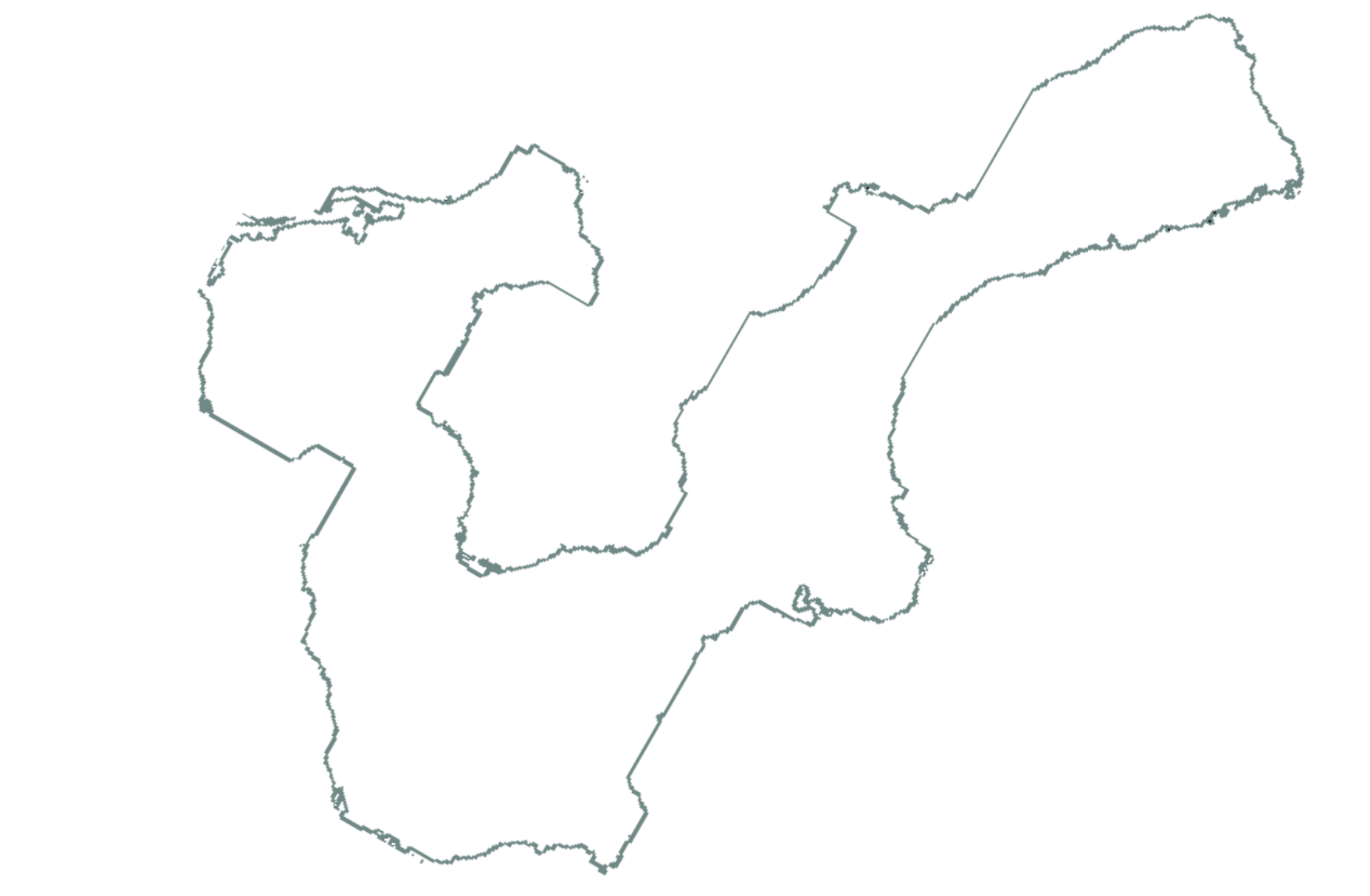}
        \caption{}
        \label{fig:2D-map}
    \end{subfigure}
    \hfill
    \begin{subfigure}[b]{0.46\linewidth}
        \centering
        \includegraphics[width=\linewidth]{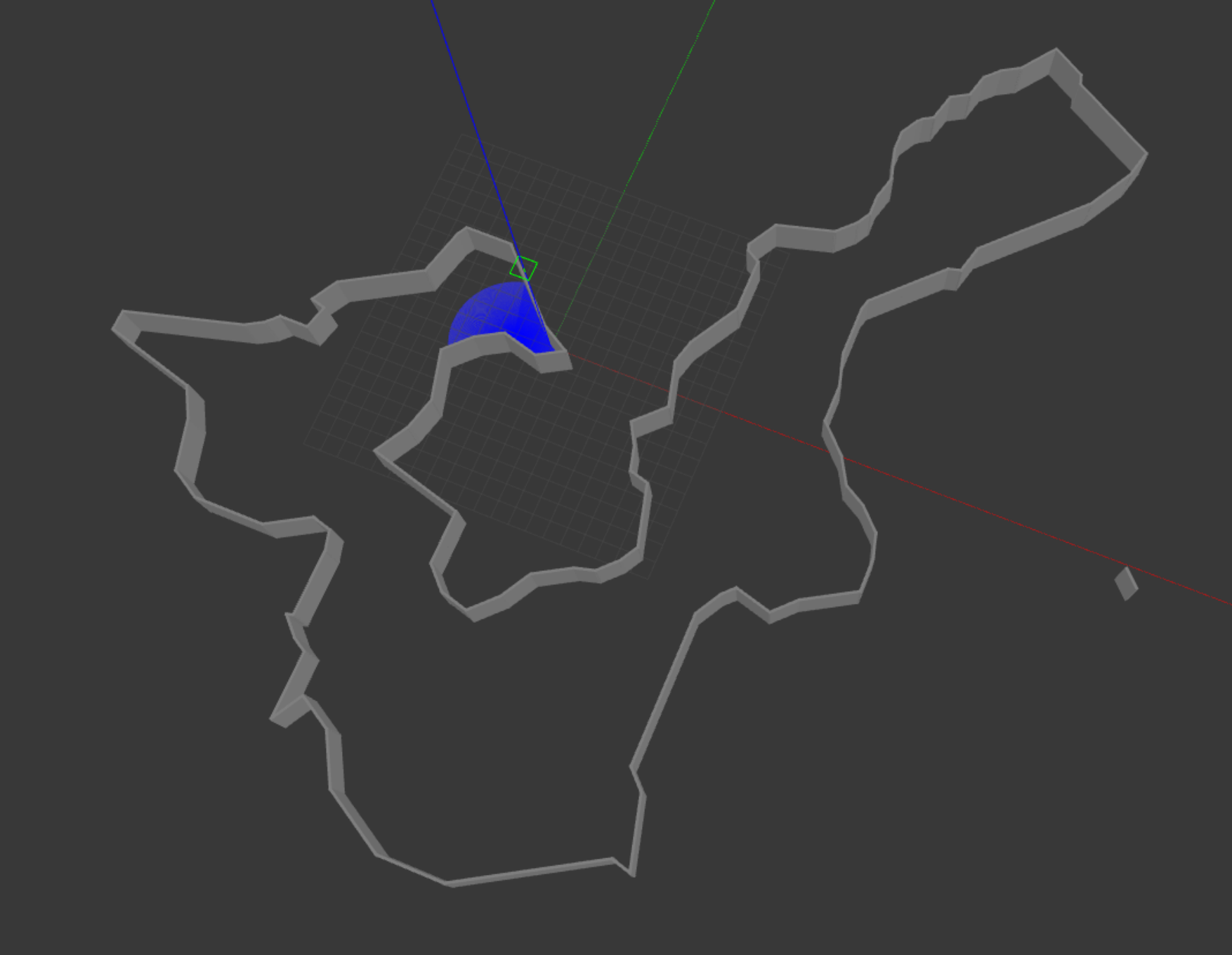}
        \caption{}
        \label{fig:3D-map}
    \end{subfigure}
    
    \begin{subfigure}[b]{\linewidth}
        \centering
        \includegraphics[width=\linewidth]{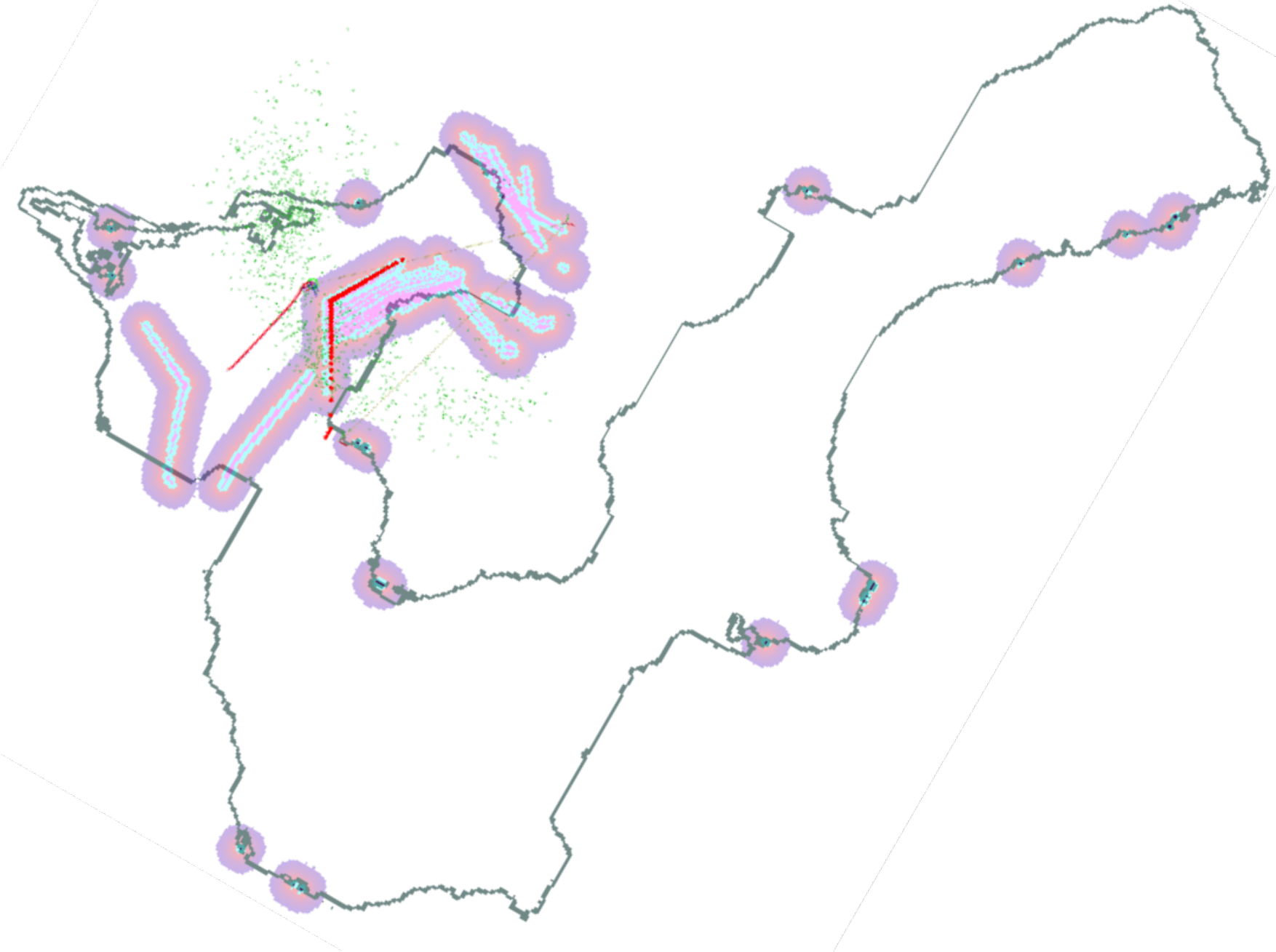}
        \caption{}
        \label{fig:path}
    \end{subfigure}
    \caption{(A) and (B) show 2D and 3D maps of Achocalla Lagoon, respectively. (C) Visualization of the global and local navigation layers in RViz during simulation in Gazebo using the Smac Hybrid A* algorithm. The cyan area represents the global costmap (static map from Gazebo), while the purple area corresponds to the local costmap (dynamic layer updated with sensor data). The red line indicates the optimal trajectory computed by the Smac Hybrid A* planner.}
    \label{fig:2D-3D}
\end{figure}

We generated a high-resolution map of Achocalla Lagoon, La Paz, Bolivia, using satellite imagery from GeoBolivia \cite{geobolivia2025}, a national geospatial platform that centralizes and shares geographic data from Bolivian governmental institutions. The map was preprocessed for edge extraction using the Canny Edge Detection algorithm, followed by a morphological erosion operation, resulting in an occupancy grid (\texttt{.pgm} format). The extracted edges captured both shoreline boundaries and regions of dense aquatic vegetation, which were treated as static obstacles within the navigation map. Figure~\ref{fig:2D-map} illustrates the processed occupancy grid, while Figure~\ref{fig:3D-map} shows its corresponding 3D representation within the Gazebo environment used for navigation validation. This navigation setup was visualized and monitored in RViz, the 3D visualization tool in ROS for sensor and robot state data, while Gazebo simulations validated map alignment, localization accuracy, and path execution before field deployment.

For position estimation, the \texttt{navsat} stack was configured to fuse GPS and IMU measurements using an Extended Kalman Filter (EKF) \cite{thrun2005, moore2014}, providing accurate estimates of the USV's displacement and heading. The IMU data are used in the prediction (time-update) stage to propagate the vehicle's state, while the GPS data are incorporated in the correction (measurement-update) stage to bound the accumulated drift. The general EKF fusion equations are given by:

\begin{equation}
\begin{split}
K_k &= P_k^{-} H_k^\top \big(H_k P_k^{-} H_k^\top + R_k\big)^{-1}, \\[2mm]
\hat{x}_k^{+} &= \hat{x}_k^{-} +
                K_k\big(z_k - h(\hat{x}_k^{-})\big), \\[2mm]
P_k^{+} &= (I - K_k H_k) P_k^{-}
\end{split}
\label{eq:ekf_fusion}
\end{equation}

where $\hat{x}{k}^{-}$ and $P{k}^{-}$ denote the predicted state and its covariance (obtained from IMU propagation), $z_k$ is the measurement vector (GPS position), $H_k$ is the Jacobian of the observation model $h(\cdot)$, $R_k$ is the measurement noise covariance, and $K_k$ is the Kalman gain, which determines the optimal blending of the two information sources. This formulation provides a minimum-variance estimate under the assumption of Gaussian noise, effectively fusing the inertial and positional data into a consistent state estimate of the USV's pose and velocity. In practice, the implementation corresponds directly with the ROS \texttt{robot\_localization} package, where the IMU topic feeds the prediction stage and the GPS topic provides the correction updates.

\subsubsection{Global path planning}
Global path planning was based on the Smac Hybrid-A* algorithm \cite{macenski2022}, developed by Steve Macenski as part of the ROS 2 Nav2 framework. This planner builds upon the original Hybrid-A* algorithm proposed by Dolgov et al. \cite{dolgov2008}, extending the conventional A* search into continuous space. Unlike standard grid-based planners, the Smac Hybrid-A* expands the search space to $(x, y, \theta)$, where the vehicle’s orientation $\theta$ is explicitly considered. This feature makes the planner suitable for nonholonomic platforms, including the USV, whose motion is constrained by the turning radius and cannot be arbitrarily discretized on a 2D grid, as illustrated in Figure \ref{fig:path}. 

The Smac Hybrid-A* algorithm can also be represented mathematically as follows. The algorithm searches for an optimal trajectory $\tau^*$ that minimizes a composite cost function $J(\tau)$, defined as:
\begin{equation}
\tau^* = \arg\min_{\tau \in \mathcal{T}} J(\tau) =
\arg\min_{\tau \in \mathcal{T}} \sum_{i=1}^{N}
\left( w_d, d_i + w_\kappa, \kappa_i^2 + w_s, (\Delta \theta_i)^2 \right)
\label{eq:cost_function}
\end{equation}
where $d_i$ represents the Euclidean distance from the current state to the nearest obstacle, $\kappa_i$ denotes the local curvature, and $\Delta \theta_i$ indicates the change in heading between consecutive motion primitives. The weights
$w_d$, $w_\kappa$, and $w_s$ balance obstacle avoidance, curvature smoothness, and orientation continuity, respectively.
Successor states are generated according to the kinematic model of the USV:

\begin{equation} \begin{cases} x_{k+1} = x_k + v \cos(\theta_k)\, \Delta t, \\ y_{k+1} = y_k + v \sin(\theta_k)\, \Delta t, \\ \theta_{k+1} = \theta_k + \frac{v}{L} \tan(\delta_k)\, \Delta t, \end{cases} \label{eq:kinematic_model} \end{equation}

where $v$ is the linear velocity, $L$ is the distance between thrusters (analogous to the wheelbase in wheeled robots), and $\delta_k$ is the instantaneous steering angle. This formulation ensures that the generated trajectories are dynamically feasible and consistent with the nonholonomic constraints of the platform.
The cost function thus integrates distance, curvature, and smoothness penalties to produce collision-free and executable paths. Once a feasible path is found, Nav2 applies a post-processing step to smooth the trajectory and minimize abrupt orientation changes before execution.

\subsubsection{Odometry}
\label{sec:odometry}
To enable autonomous navigation without wheel encoders, an odometry estimation method was implemented through sensor fusion between GPS and IMU data. The approach relies on an Extended Kalman Filter (EKF), shown in Equation \ref{eq:ekf_fusion}, which integrates global positioning information from the RTK GPS with orientation and inertial measurements from the VectorNav VN-100. As described in Section~\ref{subsec:perception-pipeline}, both sensors are interfaced through ROS 2–native drivers, providing the USV's pose (position and heading/yaw) in real time. This information is a prerequisite for global path planning algorithms, including Smac Hybrid-A*, which require the robot's current pose to compute feasible navigation trajectories.

\begin{figure}[H]
\centering
\includegraphics[width=\linewidth]{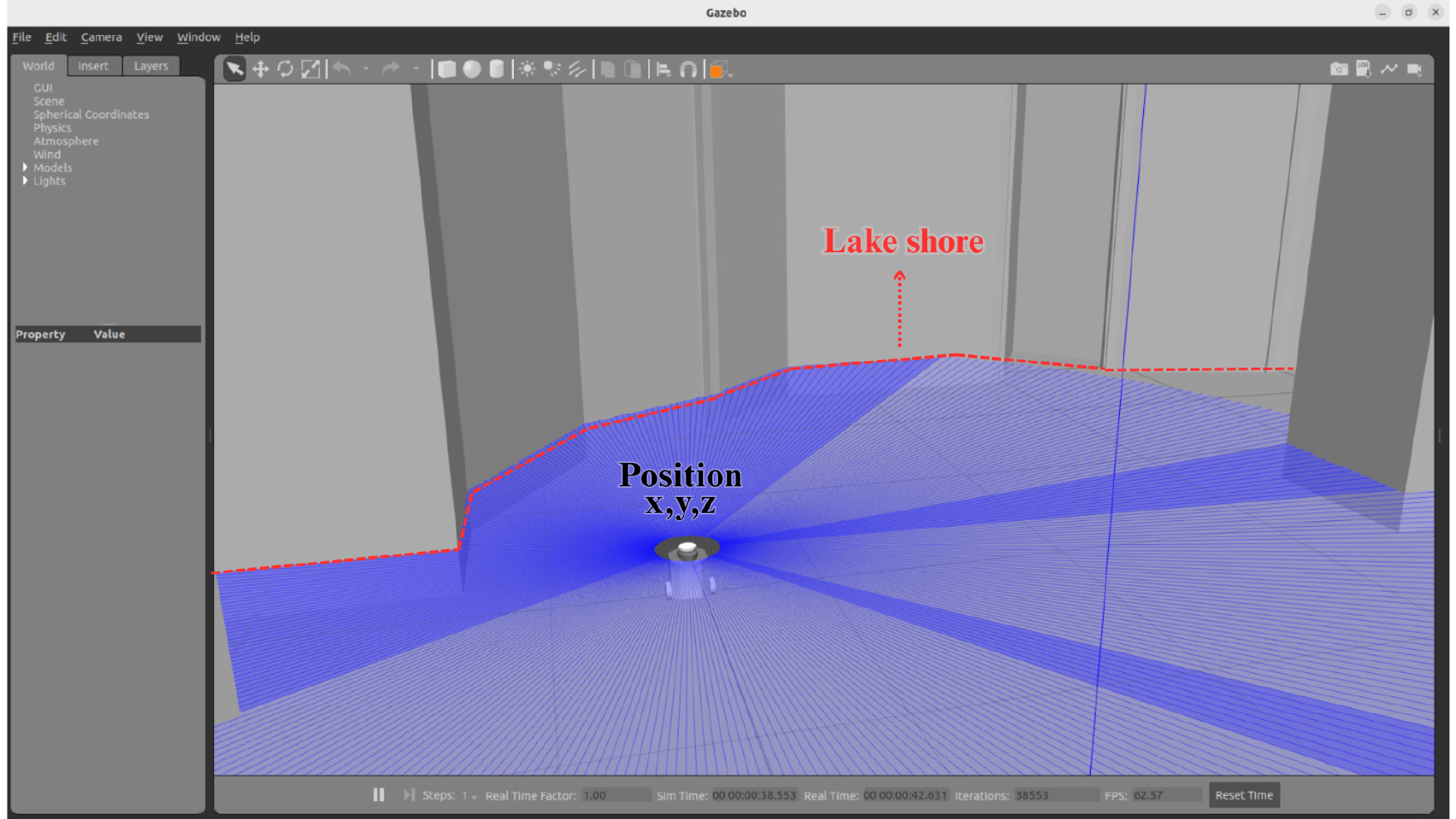}
\caption{Odometry simulation in Gazebo using the TurtleBot3 package. The scenario emulates aquatic navigation, evaluating the Extended Kalman Filter’s performance in estimating the robot’s pose from GPS and IMU data without wheel encoders.}
\label{fig:odometry-gazebo}
\end{figure}

To validate the odometry framework prior to field deployment, simulations were conducted in Gazebo using the TurtleBot3 model. Reference paths for the tests were generated with an A*-based planner, while the local planner executed the path in closed loop. Sensor models in Gazebo used plugins for GNSS and IMU and included additive zero-mean Gaussian noise and configurable standard deviations; additionally, external velocity disturbances (combination of low-frequency drift and higher-frequency white noise) were injected to emulate wind/current effects typical of aquatic environments. The simulation, shown in Figure~\ref{fig:odometry-gazebo}, confirmed that the filter's output was sufficiently stable and accurate for closed-loop path planning and execution.

\subsubsection{Obstacle detection and avoidance}

Obstacle avoidance is performed using a stereo depth estimation pipeline based on the Intel RealSense D435i camera. This sensor generates synchronized RGB and depth data using its dual infrared modules, enabling dense 3D point cloud reconstruction of the environment ahead of the USV. The \texttt{ros2\_realsense} driver manages data acquisition and publishes depth images and structured point clouds in real time, providing the perception layer with continuous spatial information.

The pipeline defines a region of interest (ROI) within the forward-facing camera view, focusing computation on the navigational corridor ahead. Depth measurements within this ROI are evaluated against a fixed threshold distance: if an object is detected within this range, it is flagged as an obstacle. This information is used to trigger path replanning or local maneuvering to avoid collisions. Compared to single-point sensing modalities such as LiDAR or sonar, this stereo vision approach is particularly effective for near-field perception and for detecting partially submerged or irregularly shaped obstacles. Its performance was confirmed through a series of indoor and outdoor experiments, where the D435i reliably identified objects in the 0.5–10~m range under diverse lighting conditions, validating the method’s robustness for real-world deployments.

\subsubsection{Teleoperation and remote supervision}
\label{sec:map-nav-obs}

Long-range telemetry and remote command are supported by a custom ROS 2 node developed for the LoRa SX1276 transceiver. This node serializes mission status, GPS position, motor feedback, and other relevant information, transmitting it to the ground station. In the opposite direction, remote teleoperation commands received via LoRa are republished onto internal ROS 2 topics, allowing the USV to be controlled from beyond direct radio range.

\begin{figure}[H]
\centering
\includegraphics[width=\linewidth]{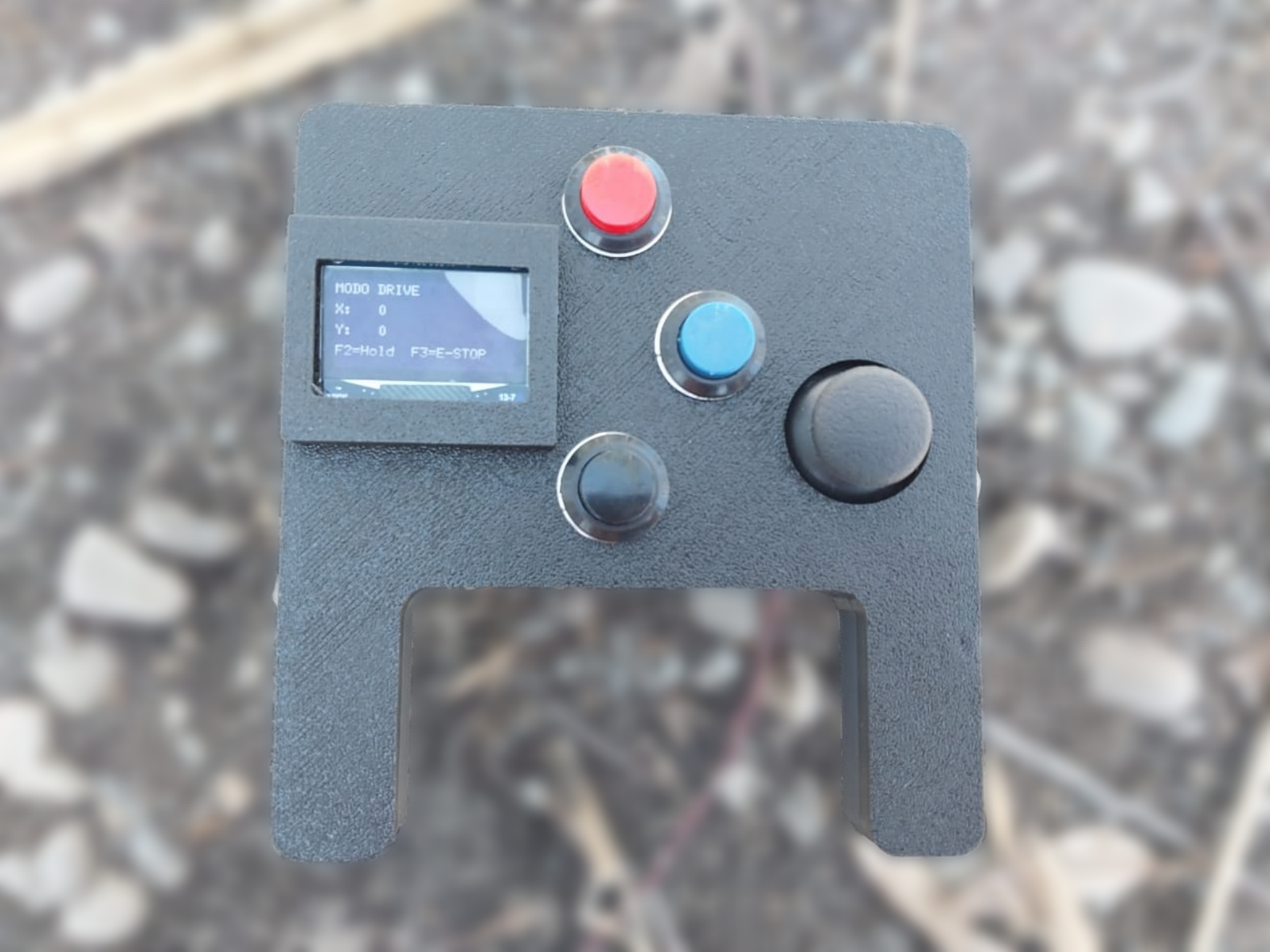}
\caption{Physical remote controller that connects to ground station for manual operation.}
\label{fig:remote_controller}
\end{figure}

A dedicated remote controller, shown in Figure~\ref{fig:remote_controller}, is designed to enable manual operation when autonomous navigation is disabled. The controller is based on an ESP32 microcontroller and integrates an OLED SSD1306 display for visualization, a dual-axis joystick, and three push buttons for manual control of propulsion and sampling modules. These inputs are transmitted via Wi-Fi using ROS 2 to a computer at the ground station, which subsequently relays the commands to the USV through the LoRa SX1276 module. This setup ensures robust and flexible bidirectional communication between the operator and the vehicle, enabling seamless transition between autonomous and manual modes.

Inside the USV, propulsion control is implemented through two independent ESP32 boards, each running a micro-ROS client. These microcontrollers subscribe to the \texttt{/cmd\_vel\_nav} topic (of type \texttt{geometry\_msgs/Twist}), receiving linear and angular velocity commands generated by the main navigation stack. The conversion from the velocity command to PWM values for each propeller follows a differential drive model, as described by:

\begin{equation}
    v_l = v_x - w_z \cdot \frac{B}{2}, \qquad
    v_r = v_x + w_z \cdot \frac{B}{2}
\end{equation}

where $v_l$ and $v_r$ are the target linear velocities for the left and right thrusters, respectively; $v_x$ is the commanded linear velocity (forward/backward motion), $w_z$ is the commanded angular velocity (yaw rate), and $B$ is the distance between the two thrusters (track width). These velocities are then mapped linearly to the corresponding PWM signals required by the ESCs, enabling coordinated control of the USV's propulsion system. The micro-ROS nodes also expose a \texttt{/motor\_status} publisher for system feedback and an \texttt{/emergency\_stop} service, which can be triggered autonomously or by a remote operator. 

\section{Water Sampling System}
\label{sec:water-sample-collection-mechanism}

The water sampling system is one of the principal contributions of this work, and this section provides a detailed description of its design and operation. As noted in Section~\ref{sec:literature-review}, initial tests using a pH and temperature probe (industrial BNC-type transducer, model B08VS3TDX2) together with a galvanic dissolved oxygen sensor (analog transducer, model B08JGRT7RM) in local water bodies indicated that these instruments are better suited for static water samples, as wave motion significantly affects their accuracy. Accordingly, the USV was designed to collect discrete water samples for on-shore analysis, thereby avoiding the limitations of continuous in situ measurements. The water sampling system comprises six identical square modules (see Figure~\ref{fig:sample-collection-system_A}(b) arranged in a 3×2 configuration and housed within a reinforced steel frame structure (a) integrated into the main body of the vehicle (c).

\subsection{Water sampling mechanism}

\begin{figure*}[ht]
    \centering
    \begin{subfigure}[b]{0.5\textwidth}
        \centering
        \includegraphics[width=\textwidth, trim={30 10 30 10}, clip]{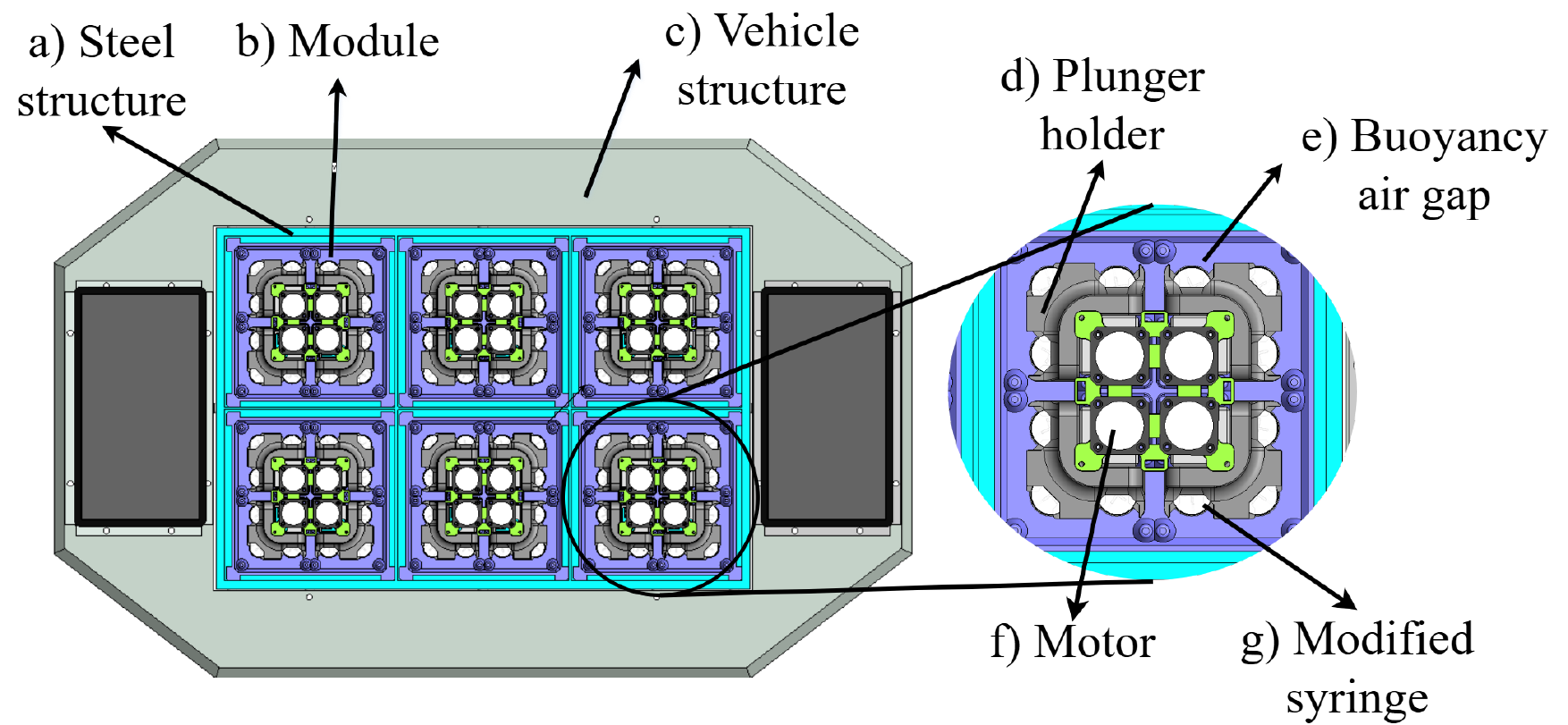}
        \caption{}
        \label{fig:sample-collection-system_A}
    \end{subfigure}
    \hfill
    \begin{subfigure}[b]{0.44\textwidth}
        \centering
        \includegraphics[width=\textwidth, trim={30 10 30 10}, clip]{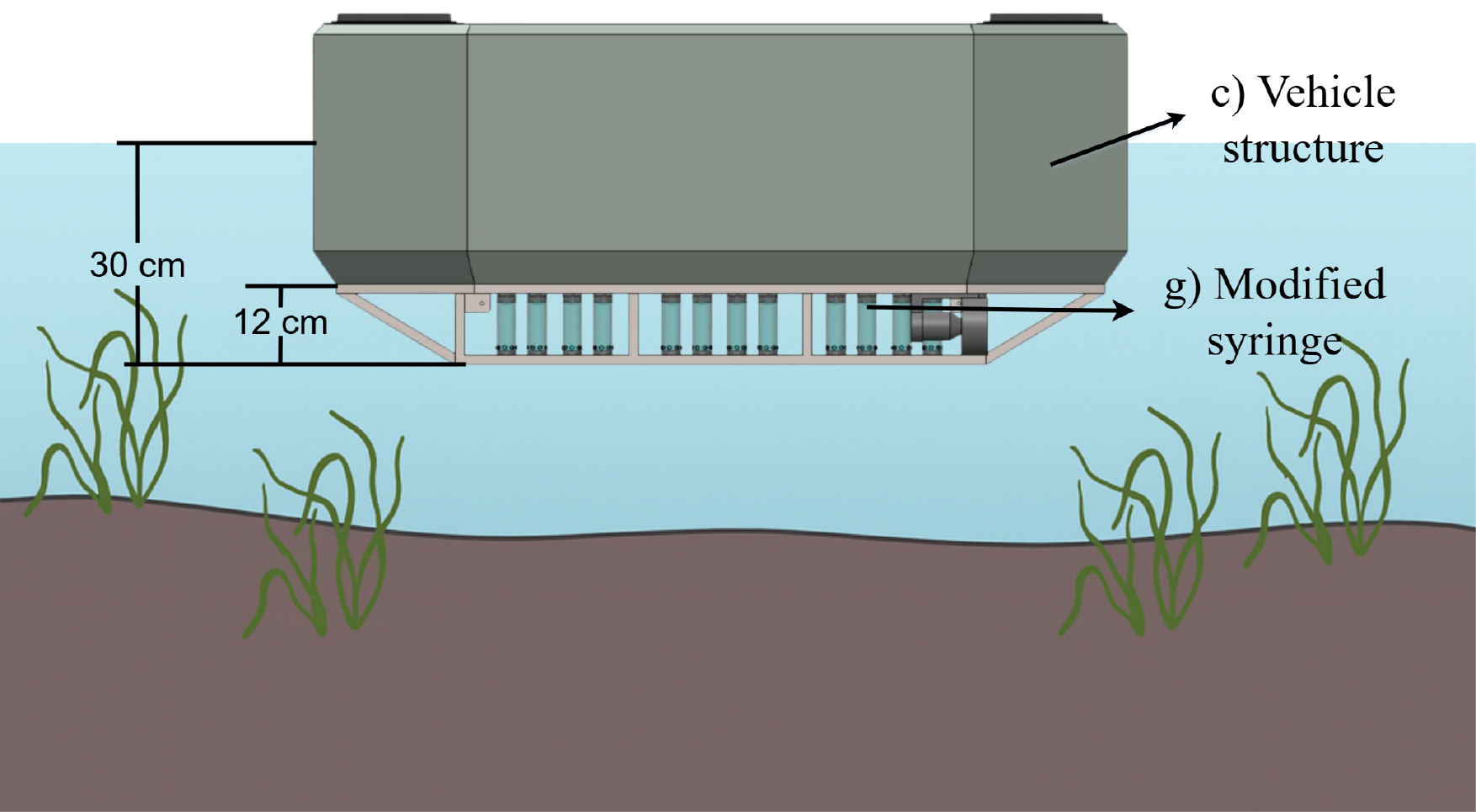}
        \caption{}
        \label{fig:sample-collection-system_B}
    \end{subfigure}

    \begin{subfigure}[b]{0.25\textwidth}
        \centering
        \includegraphics[width=\textwidth]{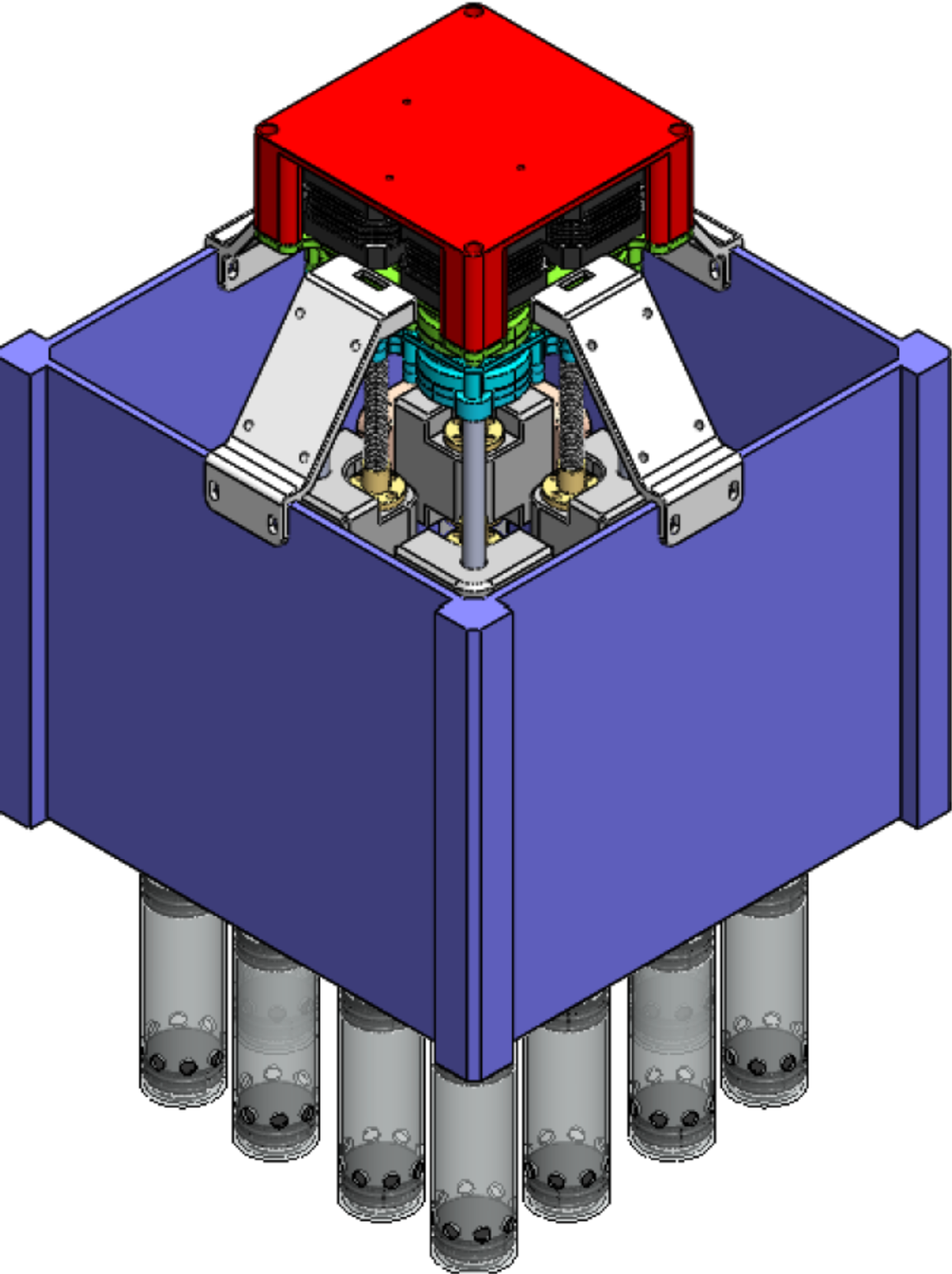}
        \caption{}
        \label{fig:sampling-module-a}
    \end{subfigure}
    \hfill
    \begin{subfigure}[b]{0.4\textwidth}
        \centering
        \includegraphics[width=\textwidth]{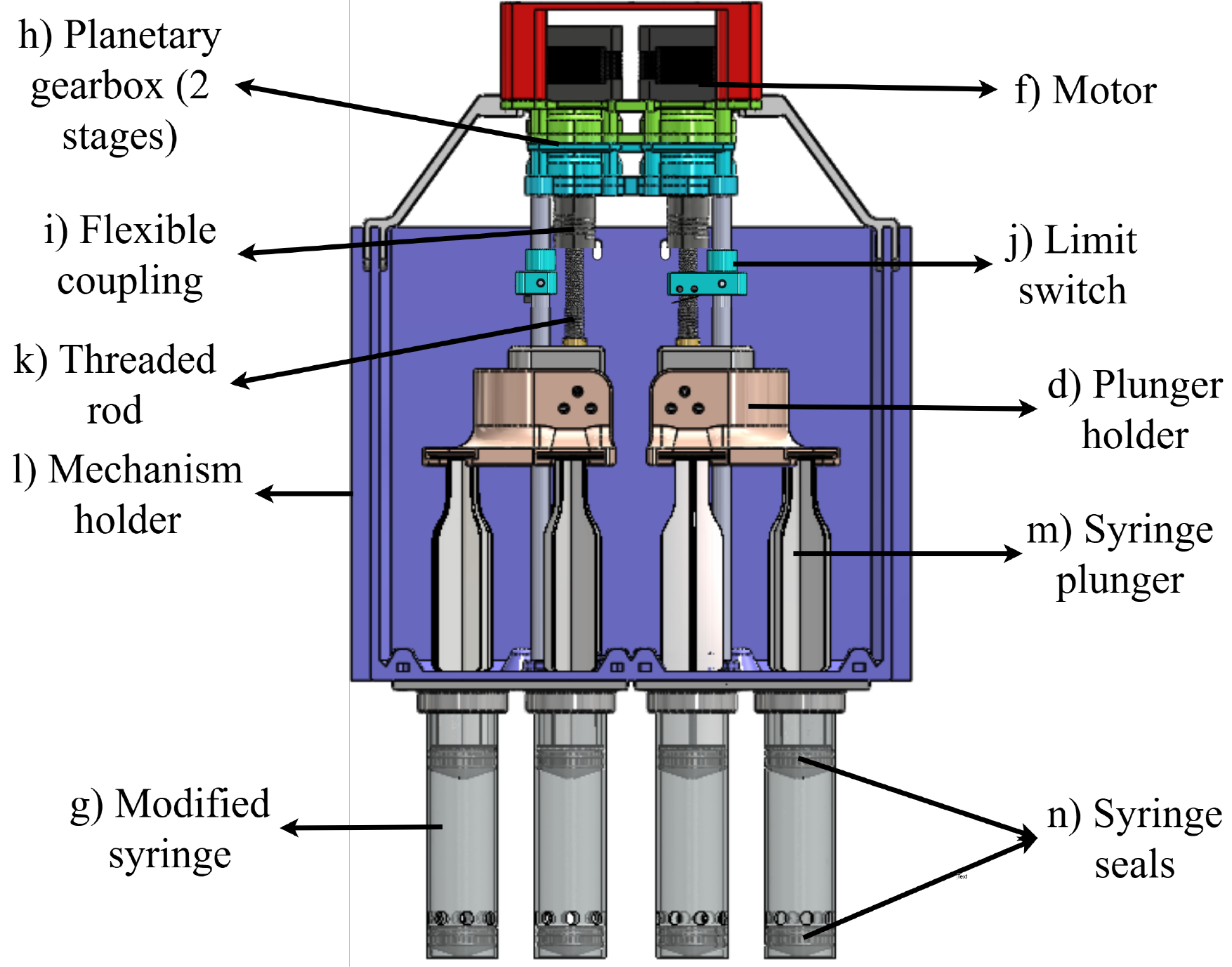}
        \caption{}
        \label{fig:sampling-module-b}
    \end{subfigure}
    \hfill
    \begin{subfigure}[b]{0.3\textwidth}
        \centering
        \includegraphics[width=\textwidth]{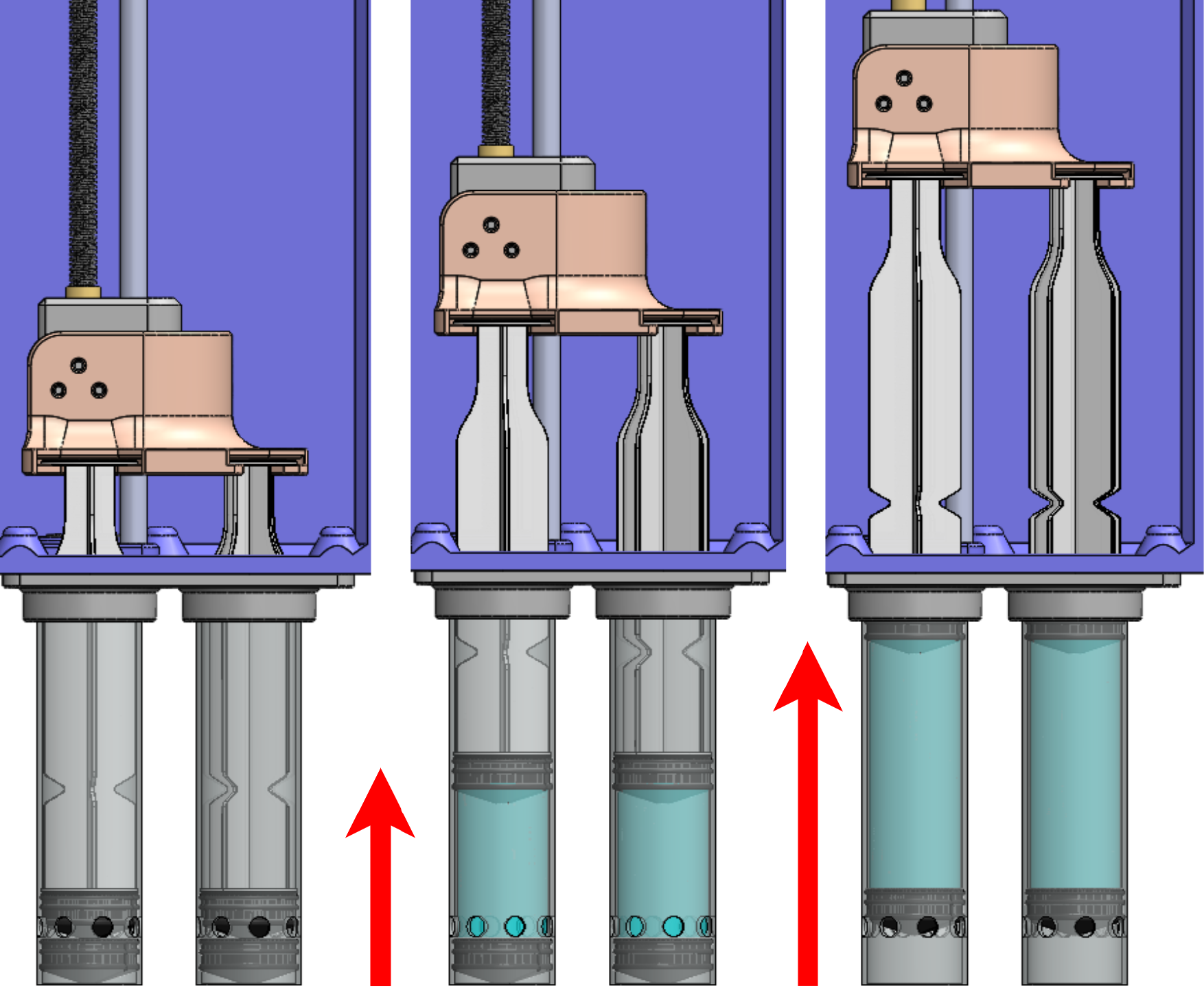}
        \caption{}
        \label{fig:sampling-module-c}
    \end{subfigure}

    \caption{(A) Top view of the water-sampling system inside the USV, including a close-up of a module with visible internal components such as motors, plunger holders, and modified syringes. (B) Side view highlighting the bottom arrangement of the syringes and the submersion depth. (C–E) Outer and inner views of a single sampling-collection module.}
    \label{fig:sample-collection-system}
\end{figure*}

Each water sampling module includes a mechanism holder (l in Figure \ref{fig:sampling-module-b}), manufactured using fused deposition modeling (FDM) additive manufacturing, with PETG as the primary material due to its mechanical strength and water resistance. Each module contains four stepper motors (f), each of which simultaneously drives three modified medical syringes (g). This configuration enables the collection of 72 individual water samples per cycle, distributed across 24 sampling points. It also facilitates the maintenance and replacement of individual units, while enhancing the system’s scalability and robustness.

Mechanically, syringe actuation is achieved through a lead-screw mechanism, similar to those used in 3D printers. Each module is equipped with four stepper motors, represented as (f) in Figure \ref{fig:sampling-module-b}, each coupled to a two-stage gearbox (h). The output of the planetary gearbox is connected to a threaded rod (k), which is coupled to the plunger holder (d) to drive a set of three syringes (g). Each rod is supported at the base by bearings to reduce friction and provide axial support. Flexible couplings (i) connect the motors to the threaded rods, allowing for misalignments. Additionally, a limit switch (j) is mounted at the top of each rod to detect the home position of the syringes.

The lead-screw system, which enables precise and reversible motion while providing reliable bidirectional control without valves, forms the core of the implemented powertrain. This actuation is driven through a two-stage mechanical reduction: 4:1 in the first stage and 3:1 in the second, resulting in a twelvefold increase in torque without significantly compromising speed. This torque amplification is essential for overcoming the combined resistance of water viscosity, syringe friction, and plunger sealing forces. Each stepper motor (NEMA 17, 0.45 Nm) drives a threaded rod via the reduction gearbox to actuate three 45 mL modified syringes simultaneously. The total output torque, $\tau_{\text{out}}$, is given in Equation~\ref{eq:torque}, and the system completes each aspiration–expulsion cycle in approximately 90 seconds, corresponding to an average flow rate per syringe, $Q$, defined in Equation \ref{eq:avg-flow-rate}. This value is calculated from the full displacement of the syringes over time $t$, assuming uniform motion, where $V_{\text{total}}$ is the total volume displaced by a syringe.

\begin{equation}
\tau_{\text{out}} = 12 \cdot \tau_{\text{motor}} = 12 \cdot 0.45\ \text{Nm} = 5.4\ \text{Nm}
\label{eq:torque}
\end{equation}
\begin{equation}
Q = \frac{V_{\text{total}}}{t} = \frac{45\ \text{mL}}{90\ \text{s}} = 0.5\ \text{mL/s}
\label{eq:avg-flow-rate}
\end{equation}

\begin{figure}[ht]
\centering
\includegraphics[width=0.7\linewidth]{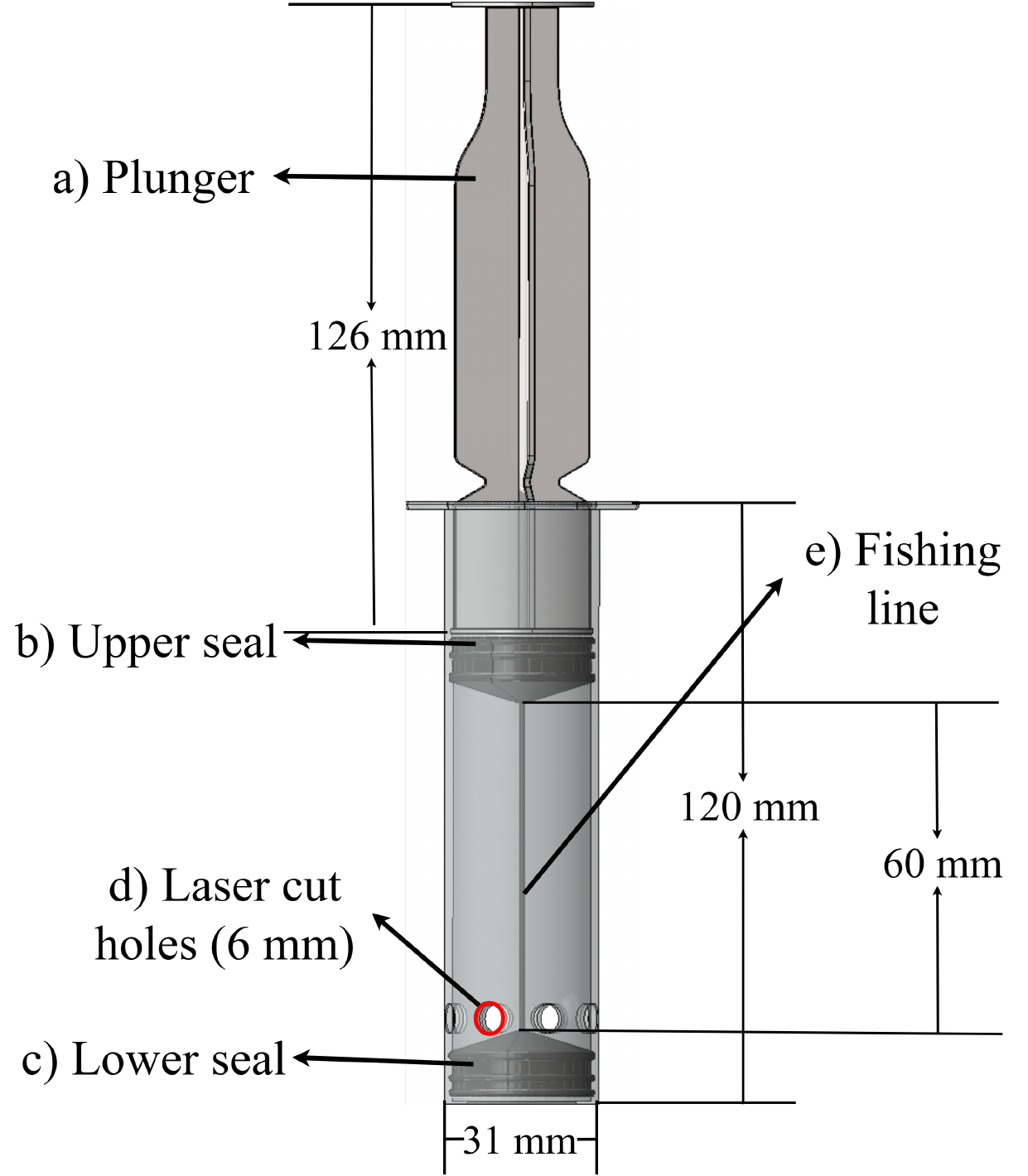}
\caption{Modified medical syringe.}
\label{fig:modified-syringe}
\end{figure}

\begin{figure*}[ht]
    \centering
    \begin{subfigure}[b]{0.58\textwidth}
        \centering
        \includegraphics[width=\textwidth]{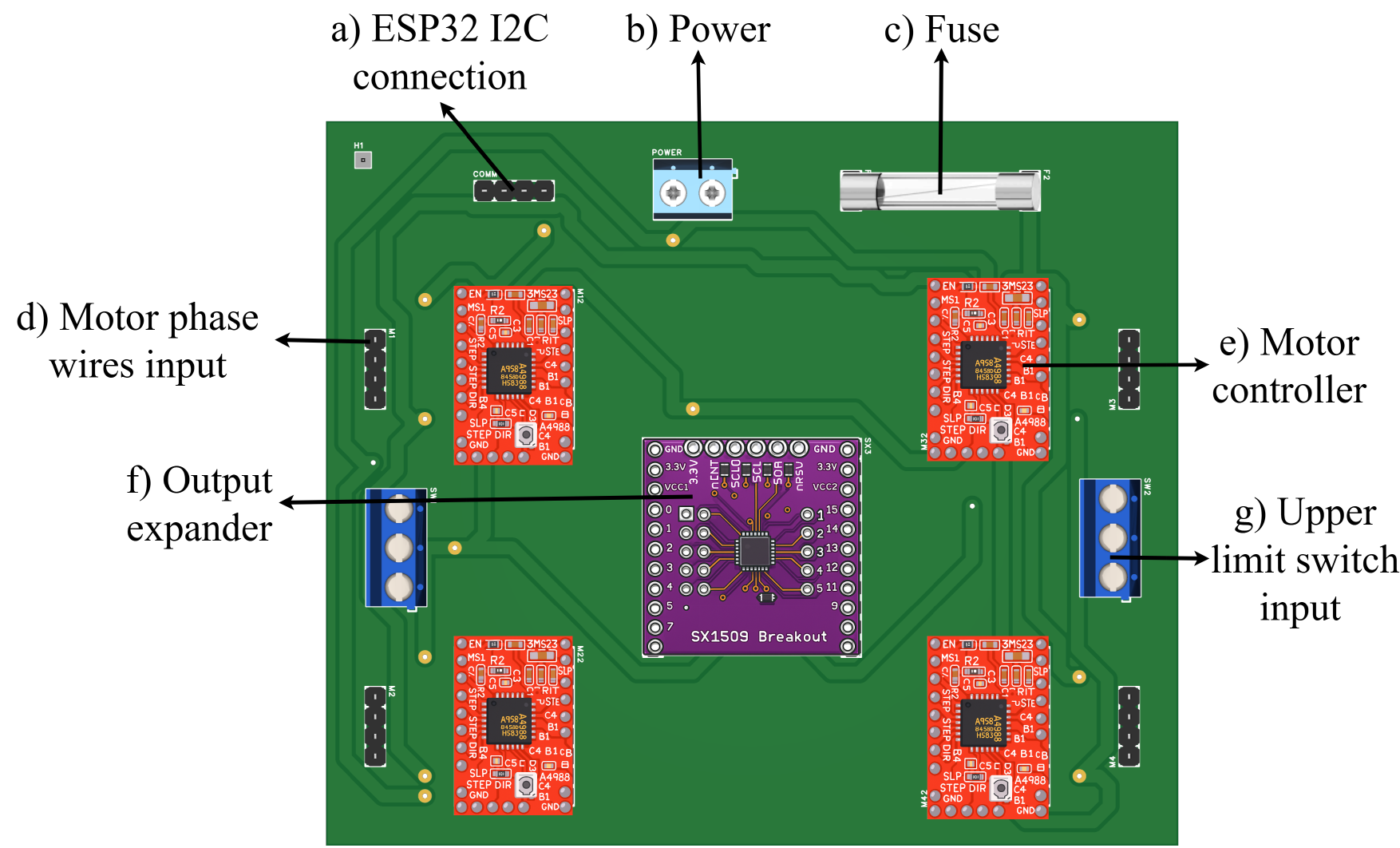}
        \caption{}
        \label{fig:pcb-collection-mech_1}
    \end{subfigure}
    \hfill
    \begin{subfigure}[b]{0.36\textwidth}
        \centering
        \includegraphics[width=\textwidth]{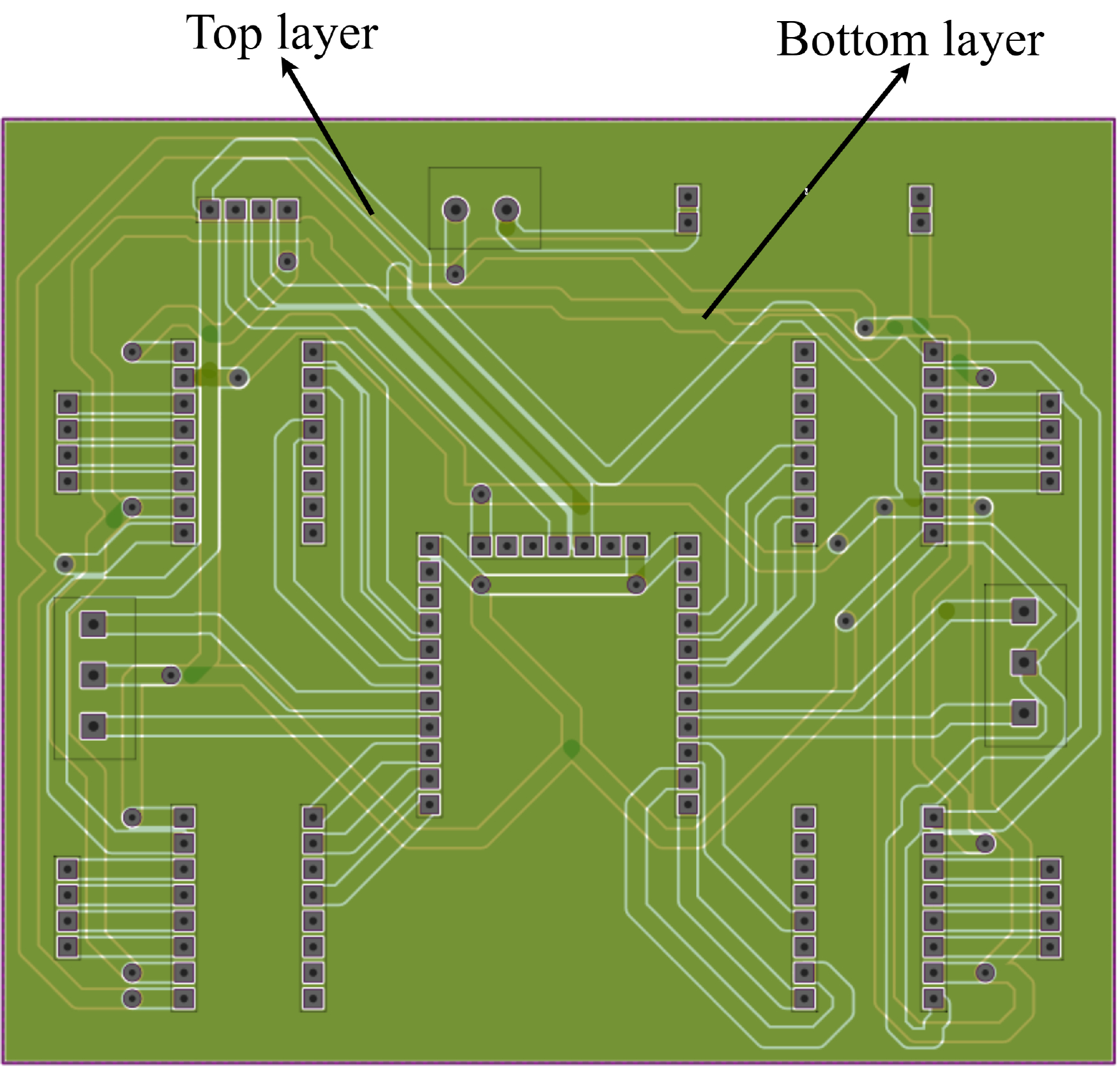}
        \caption{}
        \label{fig:pcb-collection-mech_2}
    \end{subfigure}
    \vspace{0.52cm} 
    \begin{subfigure}[b]{\textwidth}
        \centering
        \includegraphics[width=0.95\textwidth]{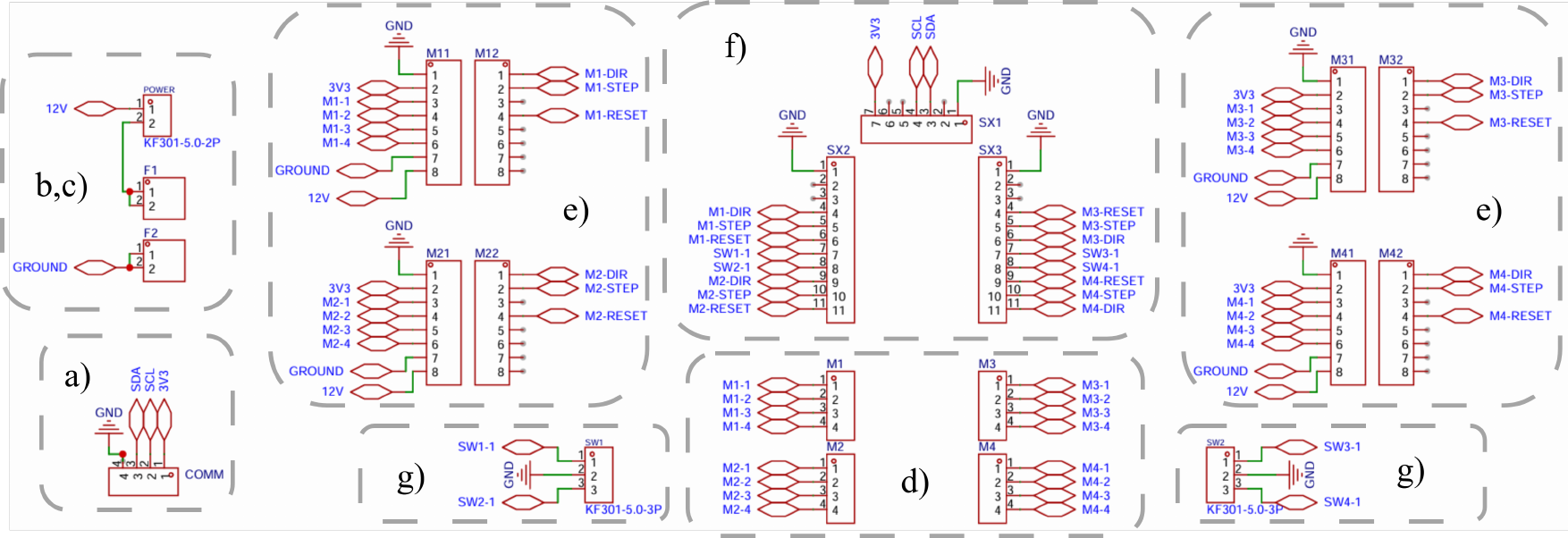}
        \caption{}
        \label{fig:pcb-collection-mech_3}
    \end{subfigure}

    \caption{A) shows a top view, including the electronic components such as the output expander, motor controller and connection inputs. B) presents an internal PCB layout view and C) illustrates the complete schematic of the PCB, detailing each circuit block}
    \label{fig:pcb-collection-mech}
\end{figure*}

To prevent leakage and cross-contamination, each syringe was mechanically modified; see Figure \ref{fig:modified-syringe}. The plunger (a) remains unchanged and is actuated via the screw mechanism to perform suction. A custom double-sealing system was implemented using two independent seals. The upper seal (b) maintains vacuum integrity during the aspiration stage, while the lower seal (c) isolates the stored sample once the plunger returns to its top position. To facilitate water intake, six lateral holes with a diameter of 6~mm were laser-cut into the syringe wall using computer numerical control (CNC) equipment, allowing efficient water entry during immersion. A fishing line (e), placed along the axis of the syringe and attached to the seals, enables controlled actuation of the lower seal by pulling it upward during the final stage of the sampling cycle, as shown in Figure~\ref{fig:sampling-module-c}. When the plunger reaches the top position, the fishing line applies tension to the lower seal, drawing it against the base of the syringe to fully isolate the sample. 

For the electronics of the sample collection modules, we used an SX1509 I/O expander to control four motor drivers and the upper limit switches. This configuration enables precise management of the syringes' upward and downward motion, as well as detection of the sampling cycle’s completion. An ESP32 microcontroller communicates with the SX1509 via the I\textsuperscript{2}C protocol, while the expander delivers individual control signals to each motor driver. The PCB (Printed Circuit Board) integrates all components, including the ESP32 I\textsuperscript{2}C connection (a), power supply input (b), fuse protection (c), motor driver sockets (e), motor phase wire inputs (d), the SX1509 output expander (f), and connectors for the upper limit switches (g). In addition, the PCB layout illustrates the routing of conductive tracks across the top and bottom layers, while the complete schematic diagram details the interconnection of each circuit block and signal flow. These views are presented in Figure~\ref{fig:pcb-collection-mech}.

\begin{figure*}[ht]
\centering
\begin{subfigure}[b]{0.455\linewidth}
    \centering
    \includegraphics[width=\linewidth]{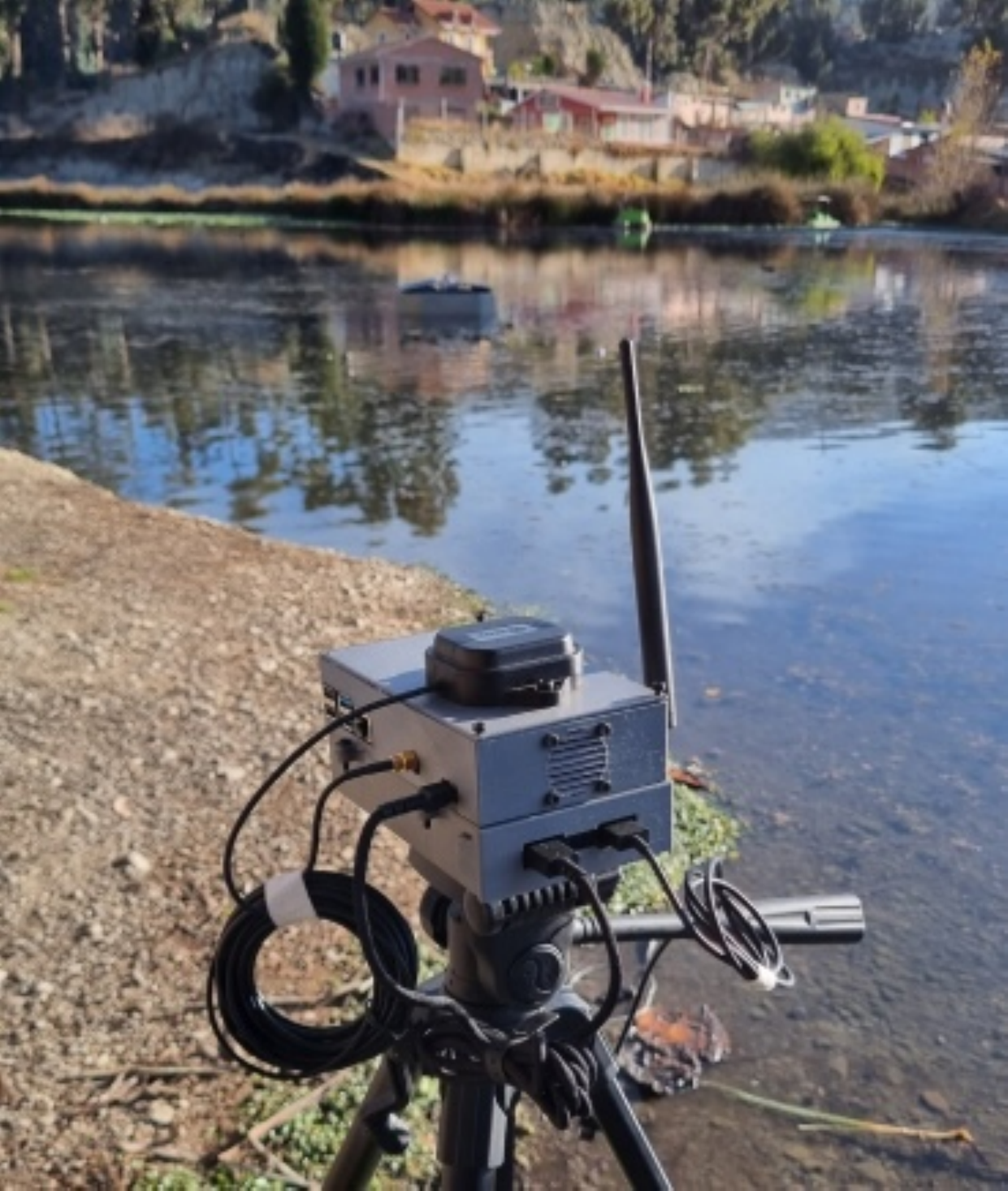}
    \caption{}
    \label{fig:data-collection-terminal}
\end{subfigure}
\hfill
\begin{subfigure}[b]{0.48\linewidth}
    \centering
    \fbox{\includegraphics[width=\linewidth]{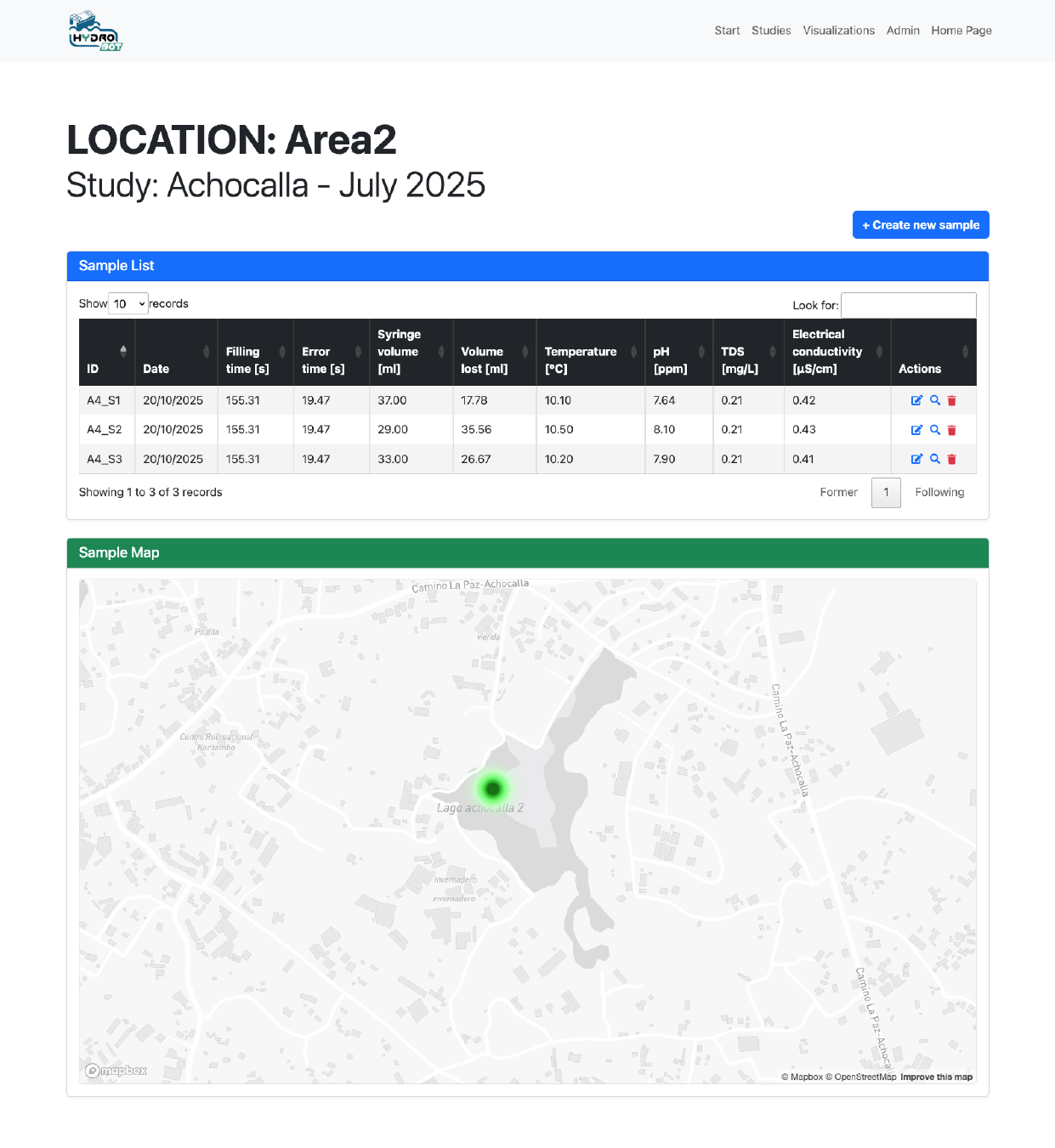}}
    \caption{}
    \label{fig:web-interface}
\end{subfigure}

\caption{Data collection system comprising: (A) the 3D-printed terminal used for in-field sampling, and (B) the synchronised web application deployed both locally on the terminal and on DigitalOcean.}
\label{fig:data-collection-combined}
\end{figure*}

\subsection{Sampling modules control}

The water sampling modules are managed by distributed ESP32 microcontrollers running micro-ROS, which integrate into the main ROS 2 network. Each ESP32 interfaces with four SX1509 I/O expanders, allowing precise and independent control of all stepper motors across the sampling units. Commands are received through the \texttt{/motor\_command} topic (\texttt{std\_msgs/UInt8MultiArray}), which encodes the target module, motor, and action (stop, forward, or reverse). Upon receiving these commands, the microcontroller activates or changes the motor direction accordingly, ensuring synchronized operation.

Each module transmits its operational state through the \texttt{/motor\_status} topic (\texttt{std\_msgs/String}), reporting parameters such as motor activity, direction, and home switch feedback. It works along the topics described in Section~\ref{sec:map-nav-obs}, adding the status to the same messages. Complementary diagnostic information is provided through the \texttt{/debug} topic, facilitating real-time monitoring and troubleshooting during field operation. This bidirectional communication not only enables remote supervision but also allows adaptive reconfiguration of the system if a module requires intervention or recalibration.

Safety and reliability are reinforced through the \texttt{/emergency\_stop} service (\texttt{std\_srvs/SetBool}), which can immediately disable all motors when triggered, preventing mechanical stress and unintended motion. Additionally, an automated recovery routine periodically checks the I\textsuperscript{2}C bus to detect and reinitialize any unresponsive SX1509 expanders. This distributed control architecture ensures deterministic actuation, consistent feedback, and full compatibility with higher-level ROS 2 frameworks for mission planning, logging, and teleoperation.

\begin{figure*}[ht]
    \centering
    \includegraphics[width=0.8\textwidth]{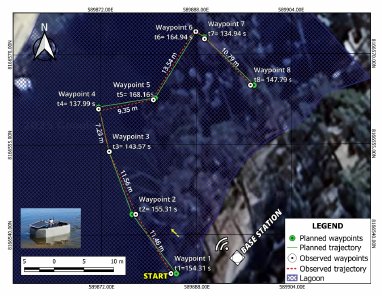}
    \caption{Location of water sampling points in Achocalla Lagoon, La Paz, Bolivia.}
    \label{fig:map}
\end{figure*}

\subsection{Data collection terminal}
\label{sec:dct}

We implemented a bespoke device that functions both as an interface for collecting and managing sample data and as a gateway for ROS 2 command transmission between the remote controller and the USV. As shown in Table \ref{tab:components}, the device consists of a Raspberry Pi 4B enclosed in a 3D-printed case and powered by a portable battery (power bank). The enclosure provides basic weather resistance and supports safe operation near the water surface. Communication with the USV is achieved through a LoRa link, enabling real-time message exchange for remote operation and status feedback, while a GPS module mounted on top supports mapping and navigation functions. Figure \ref{fig:data-collection-terminal} shows the terminal deployed in situ during the validation stage.

A Django application running on the Raspberry Pi provides local data storage to ensure operation in rural or hard-to-reach areas where connectivity is limited. Operators interact with the terminal through a web interface optimized for field conditions. A mirrored version of the same application is hosted on DigitalOcean \cite{hydrobot2025} to synchronize data and make it publicly accessible. Figure \ref{fig:web-interface} presents the application’s interface, which displays the collected data for each sampling task along with a heatmap-based map visualization.

\section{Experimental Results}
\label{sec:experimental-results}

To evaluate the USV and its sampling mechanism, four field tests were conducted at Achocalla Lagoon, La Paz, Bolivia. The first three tests were used to calibrate the system's components and address their limitations, while the final test, carried out on July 6, 2025, provided precise performance metrics that are detailed in the following subsections.

\subsection{Unmanned Surface Vehicle}

For pathfinding, we evaluated the USV's navigation and water-sampling performance across eight waypoints, as shown in Figure \ref{fig:map}. With the updated dataset, the vehicle achieved an overall waypoint precision of 87\%, with an average positional error of 0.046 m and a maximum deviation of 0.12 m from the planned trajectory. These small discrepancies were mainly influenced by propulsion motor inertia and wind disturbances. The results support the feasibility of implementing adaptive control strategies to further improve waypoint tracking under real-world operating conditions.

To monitor current consumption, we used the VictronConnect mobile application \cite{victron2025}, developed by the same company that manufactures the Smart Charge controller integrated into the USV. The application provided current and voltage measurements, allowing us to analyze power usage trends and system efficiency under different operating conditions. Specifically, the USV's two thrusters consumed approximately 1,800~W, while the Jetson Orin Nano added about 25~W, resulting in a total load of roughly 1,825~W. 

Table~\ref{tab:power} summarizes the power distribution across all onboard components, based on the specifications in Table~2 and the active-duty cycle of the sampling system. In addition to propulsion and onboard computation, the vehicle continuously powers two ESP32 microcontrollers and the sensing/communication suite, while the 24 Nema17 stepper motors used for sampling operate sequentially across different waypoints.

\begin{table}[ht]
\centering
\caption{Average electrical power consumption during a representative mission.}
\label{tab:power}
\begin{tabular}{lcc}
\toprule
\textbf{Subsystem} & \textbf{Power (W)} & \textbf{Notes} \\
\midrule
Thrusters (2 × 900~W) & 1800 & Continuous \\
\makecell[l]{Jetson Orin Nano \\(25~W mode)} & 25 & Continuous \\
ESP32 (2 × 1.65~W) & 3.3 & Continuous \\
Sensors + Comms & 4.65 & Continuous \\
\makecell[l]{Nema17 motors \\(24 × 20.4~W × 10\%)} & 48.96 & \makecell{Sequential \\operation} \\
\midrule
Total average power & 1882 & W (1.88~kW) \\
\bottomrule
\end{tabular}
\end{table}

Using the total power in Table~\ref{tab:power}, the theoretical battery endurance can be obtained from:

\begin{equation}
t_{\mathrm{bat}}
= \frac{E_{\mathrm{use}}}{\bar{P}_{\mathrm{tot}}}
= \frac{1920~\mathrm{Wh}}{1882~\mathrm{W}}
\approx 1.02~\mathrm{h}
\approx 61~\mathrm{min}.
\label{eq:battery_endurance}
\end{equation}

\begin{figure}[ht]
  \centering
  \begin{subfigure}[t]{0.48\columnwidth}
    \centering
    \includegraphics[width=\linewidth]{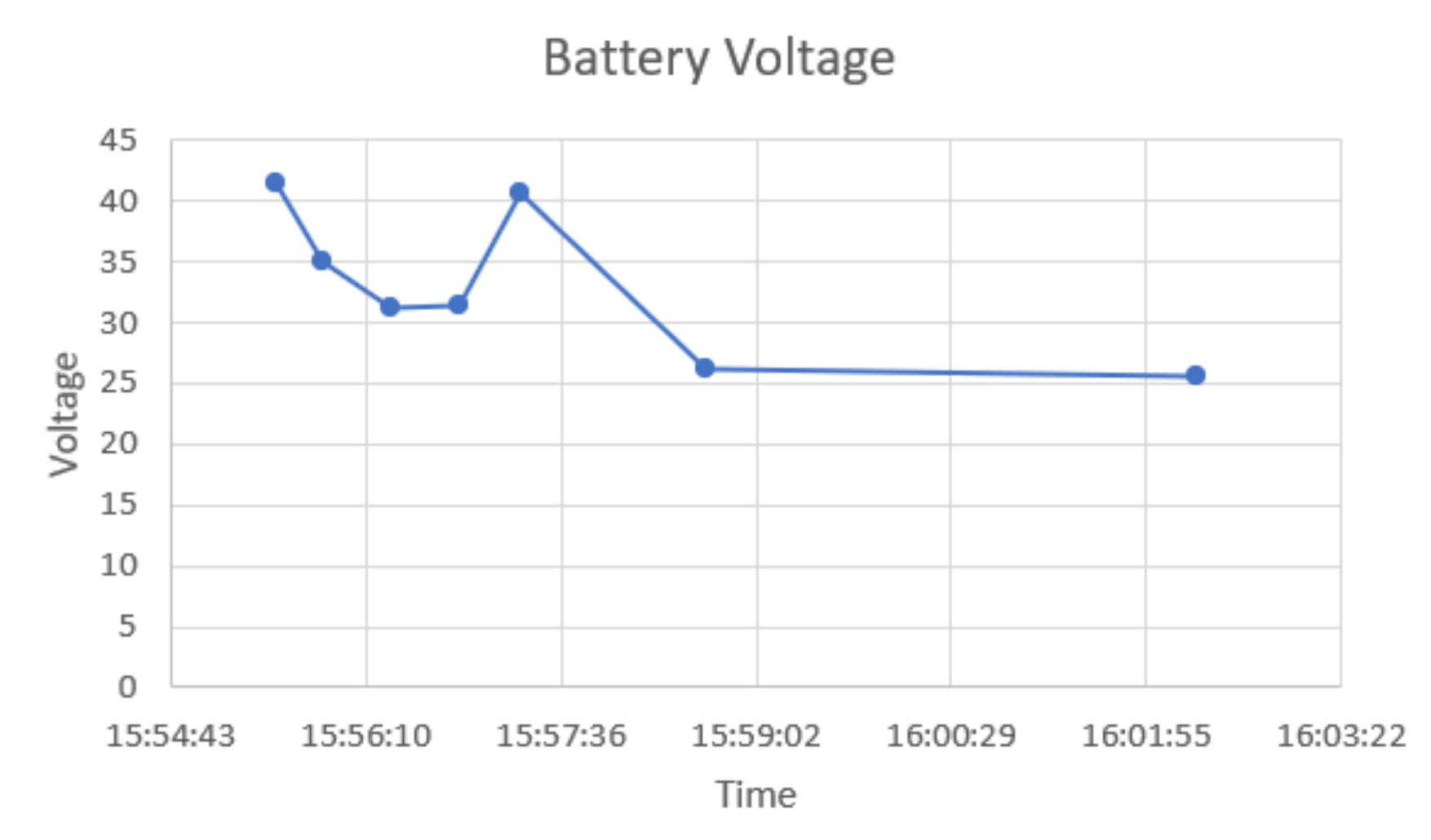}
    \caption{}
    \label{fig:sub1}
  \end{subfigure}
  \hspace{0.002\columnwidth}
  \begin{subfigure}[t]{0.48\columnwidth}
    \centering
    \includegraphics[width=\linewidth]{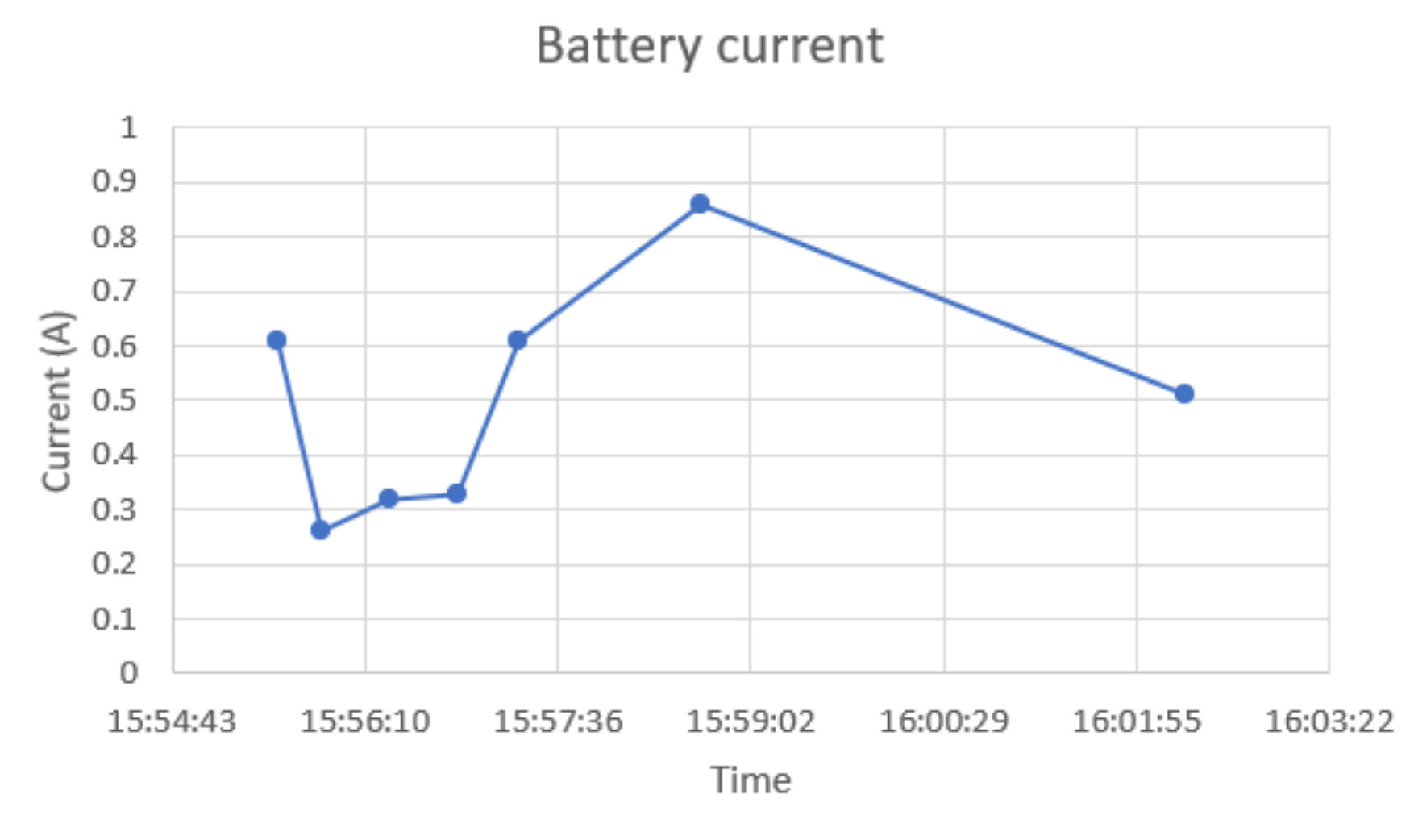}
    \caption{}
    \label{fig:sub2}
  \end{subfigure}
  \caption{Battery voltage and current readings obtained from the Victron solar charge controller via the VictronConnect mobile app.}
  \label{fig:test-routine}
\end{figure}

This value agrees closely with the approximately 60~minutes of operation recorded during field tests under high load, which increased by about 10~minutes when the solar panels were connected through the charge controller. Data collected during one of the final sampling operations, using four active modules for 10~minutes, are presented in Figures \ref{fig:sub1} and \ref{fig:sub2}. The first drop observed in Figure \ref{fig:sub2} corresponds to the energy used to move the USV toward the designated waypoint, while the second drop reflects the power consumed by the sampling motors.

Finally, the system's effective communication range was empirically evaluated by adjusting transmission parameters and conducting field trials. The maximum reliable control distance was measured at 66.8~meters, beyond which noticeable response delays were observed.

%%%%%%%%%%%%%%%%%%

\subsection{Water sampling system}

\begin{figure}[ht]
    \centering
    \includegraphics[width=\linewidth]{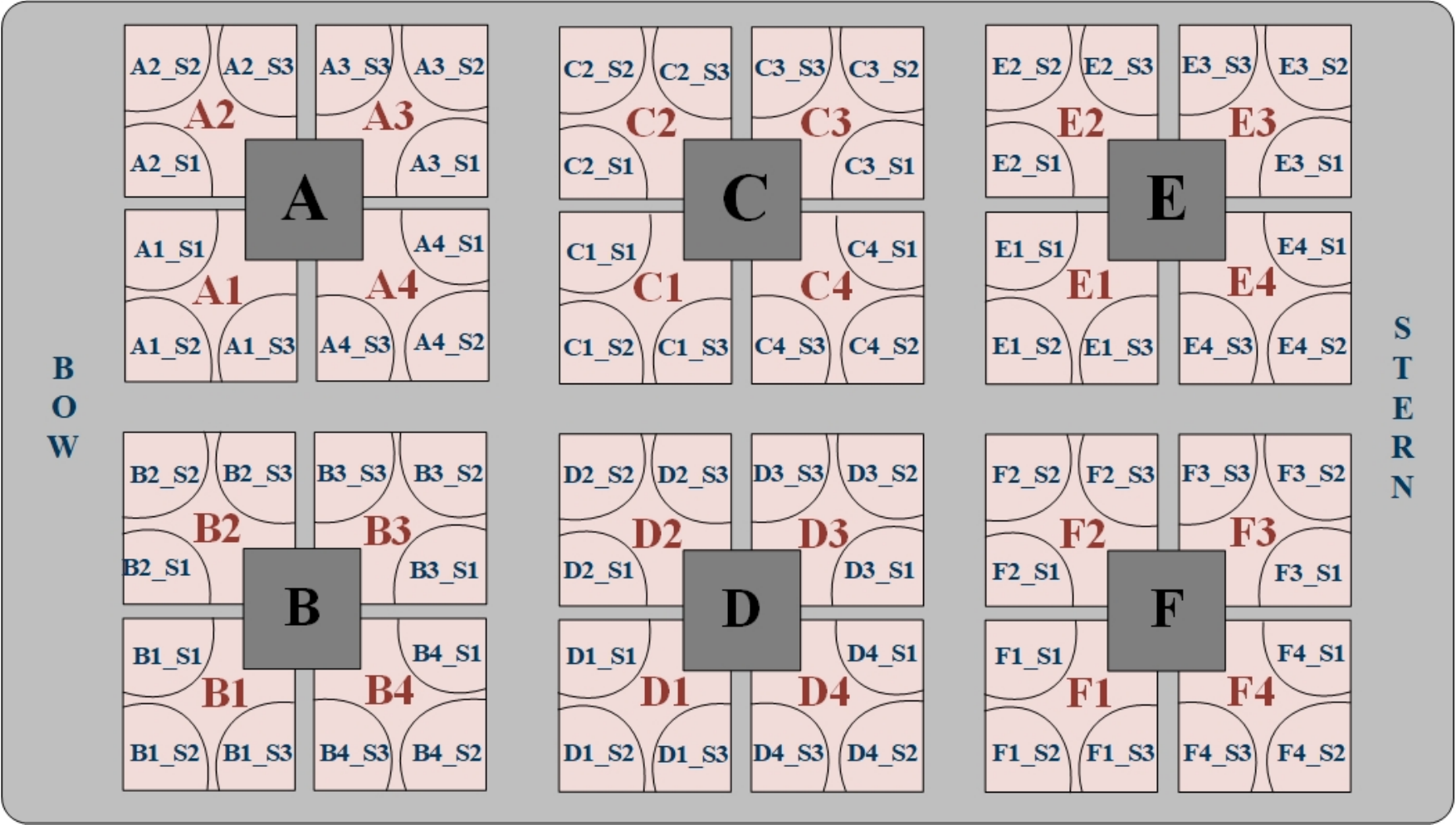}
    \caption{Labeled diagram of the water sample collection system, composed of six modules (A–F), each consisting of 12 syringe-based sampling mechanisms.}
    \label{fig:collection_system_distribution}
\end{figure}

The sampling protocol, including sample labeling, number of samples per location, and shoreline measurements performed with portable sensor equipment, followed the Binational Protocol for Water Quality Monitoring in Lake Titicaca \cite{protocolo2020} to ensure consistent data collection in shallow waters. Water was collected at a depth of 30\,cm to avoid debris and sediment, and the collection modules were labeled as shown in Figure~\ref{fig:collection_system_distribution} to facilitate sample handling.

During the final field campaign, mechanisms A3, A4, B2, B3, B4, D2, D3, and D4 successfully collected water from waypoints 1–8, respectively (see Figure \ref{fig:map}). After collection, the USV returned to shore, where temperature, pH, total dissolved solids (TDS), and electrical conductivity (EC) were measured for each sample using HANNA pH and EC/TDS testers (Table \ref{tab:table_sampling_data}). The campaign yielded an average filling time of 150.88\,s per sample, 16.06\% slower than the theoretical 130\,s, and an average collected volume of 35.25\,mL per syringe (21.67\% loss relative to the 45\,mL design capacity). Together, these metrics characterize the system’s operational performance \textit{in situ} and support subsequent analysis of performance deviations.

Individual syringes exhibited larger errors (e.g., A3\_S1: 100\% loss; D2\_S2: 46.67\% loss), identifying them as outliers that drive module-level variability. These deviations are consistent with field observations: millimeter-scale variations in hole size and fishing-line tension produced local leaks; delayed or failed activation of the limit switch and increased motor resistance due to algae or debris caused systematic increases in filling time (e.g., B4 group mean = 168.16\,s; 29.35\% error). Mechanical asymmetries in the syringe-actuation manifold further contributed to partial fills and extended cycle times. Reporting per-syringe and per-module means (and standard deviations) of volume loss and time error enables maintenance prioritization and identification of dominant failure modes.

\begin{figure}[H]
    \centering
    \includegraphics[width=\linewidth]{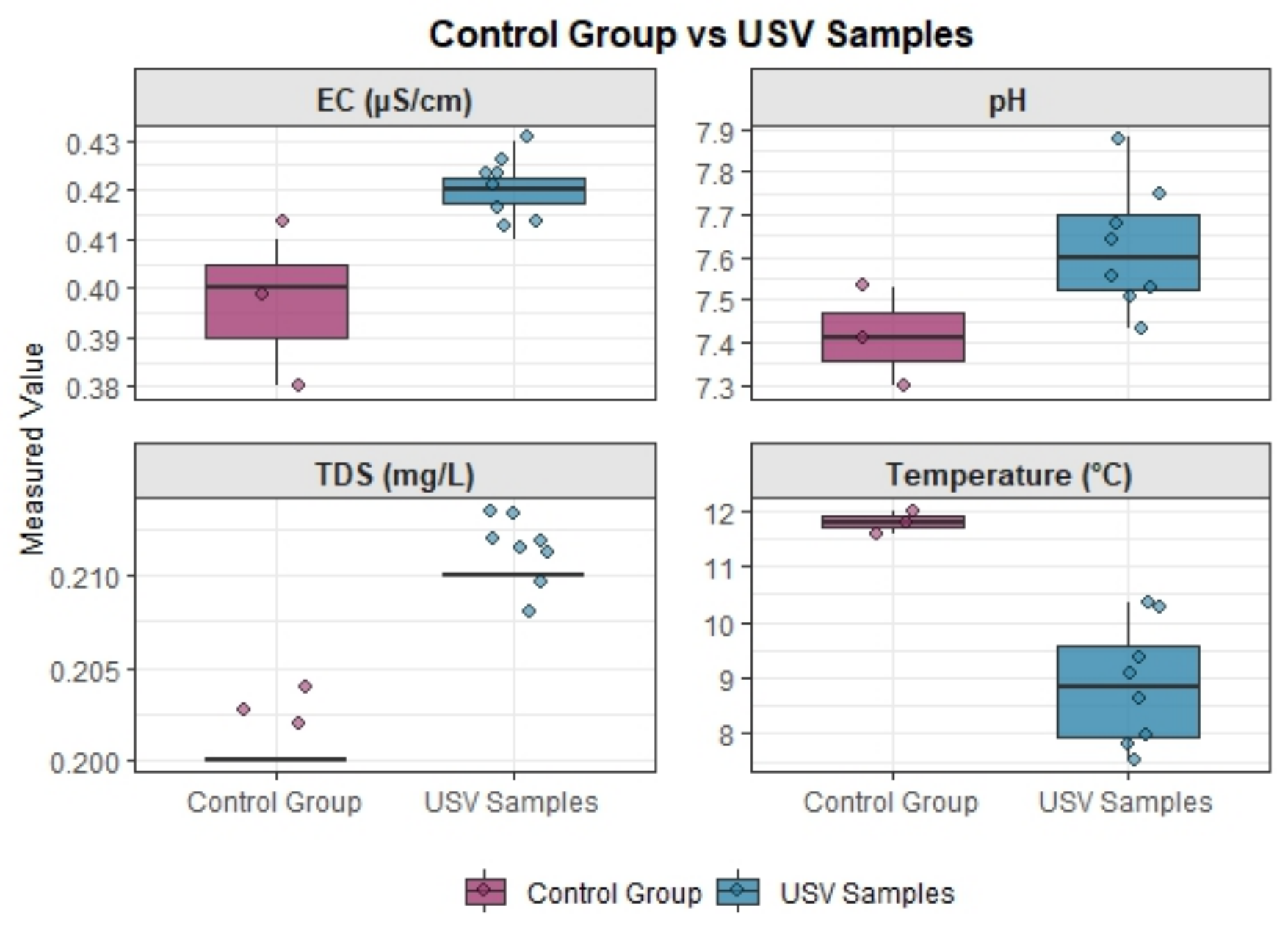}
    \caption{Comparison of in situ water quality parameters.}
    \label{fig:water-metrics-visualization}
\end{figure}

To verify that the automated sampling system did not compromise sample integrity, a necessary step in validating the USV's mechanical design, a control comparison was conducted with three manually collected shoreline samples ($N = 3$). The analysis quantified the absolute and relative errors between the mean values of the eight USV-collected samples ($N = 8$) and the control group. The parameters most sensitive to mechanical alteration, TDS and EC, showed excellent agreement, with a maximum relative error of only 5.0\% (0.01 mg/L for TDS and 0.02 µS/cm for EC). This low error confirms that the USV's syringe-based mechanism effectively preserved the primary physicochemical characteristics of the water samples, demonstrating the system’s reliability (Figure \ref{fig:water-metrics-visualization}). Although larger differences were noted in temperature (2.68 °C) and pH (0.21), these were attributed to typical diurnal cooling as sampling extended from late afternoon into nightfall. The pH difference aligns closely with historical shoreline data from the same lagoon \cite{paz2025}, which reported pH = 7.68–7.78 under similar conditions, indicating that the observed variations reflect  environmental gradients and confirming the USV's ability to capture lagoon heterogeneity accurately.

\begin{table*}[ht]
\centering
\renewcommand{\arraystretch}{0.80}

\caption{Measured water-sampling performance and in situ water-quality parameters. $\pi$ denotes the average value per column.}
\label{tab:table_sampling_data}

\newcolumntype{P}[1]{>{\centering\arraybackslash}m{#1}}

\begin{tabularx}{\textwidth}{@{} 
P{1.4cm}  % Syringe
P{1.5cm}  % Filling time
P{1.6cm}  % Time error
P{1.7cm}  % Volume
P{1.6cm}  % Volume loss
P{1.8cm}  % Temp
P{1.2cm}  % pH
P{1.5cm}  % TDS
P{1.5cm}  % EC
@{}}
\toprule
\textbf{Syringe} &
\textbf{Filling time (s)} &
\textbf{Time error (\%)} &
\textbf{Volume / syringe (ml)} &
\textbf{Volume loss (\%)} &
\textbf{Temperature (°C)} &
\textbf{pH} &
\textbf{TDS (mg/L)} &
\textbf{EC (µS/cm)} \\
\midrule

% ---------------------- A3 ----------------------
A3\_S1 & \multirow{3}{*}{154.31} & \multirow{3}{*}{18.70} & 0  & 100.00 & --    & --    & --    & --    \\
A3\_S2 &                         &                        & 39 & 13.33  & 10.60 & 7.86  & 0.20  & 0.41 \\
A3\_S3 &                         &                        & 42 & 6.67   & 10.10 & 7.41  & 0.21  & 0.43 \\
\midrule
\textbf{$\pi_{A3}$} &  & & 27 & 40 & 6.90 & 5.09 & 0.14 & 0.28 \\
\midrule

% ---------------------- A4 ----------------------
A4\_S1 & \multirow{3}{*}{155.31} & \multirow{3}{*}{19.47} & 37 & 17.78 & 10.10 & 7.64 & 0.21 & 0.42 \\
A4\_S2 &                         &                        & 29 & 35.56 & 10.50 & 8.10 & 0.21 & 0.43 \\
A4\_S3 &                         &                        & 33 & 26.67 & 10.20 & 7.90 & 0.21 & 0.41 \\
\midrule
\textbf{$\pi_{A4}$} & & & 33 & 26.67 & 10.27 & 7.88 & 0.21 & 0.42 \\
\midrule

% ---------------------- B2 ----------------------
B2\_S1 & \multirow{3}{*}{143.57} & \multirow{3}{*}{10.44} & 41 & 8.89  & 8.80 & 7.48 & 0.21 & 0.40 \\
B2\_S2 &                         &                        & 41 & 8.89  & 8.70 & 7.42 & 0.20 & 0.42 \\
B2\_S3 &                         &                        & 36 & 20.00 & 8.40 & 8.36 & 0.22 & 0.47 \\
\midrule
\textbf{$\pi_{B2}$} & & & 39.33 & 12.59 & 8.63 & 7.84 & 0.21 & 0.43 \\
\midrule

% ---------------------- B3 ----------------------
B3\_S1 & \multirow{3}{*}{137.99} & \multirow{3}{*}{6.15} & 42 & 6.67  & 9.40 & 7.67 & 0.21 & 0.41 \\
B3\_S2 &                         &                       & 38 & 15.56 & 9.30 & 7.58 & 0.21 & 0.41 \\
B3\_S3 &                         &                       & 38 & 15.56 & 9.40 & 7.39 & 0.20 & 0.42 \\
\midrule
\textbf{$\pi_{B3}$} & & & 39.33 & 12.60 & 9.37 & 7.55 & 0.21 & 0.41 \\
\midrule

% ---------------------- B4 ----------------------
B4\_S1 & \multirow{3}{*}{168.16} & \multirow{3}{*}{29.35} & 33 & 26.67 & 8.90 & 7.46 & 0.21 & 0.42 \\
B4\_S2 &                         &                        & 34 & 24.44 & 9.10 & 7.41 & 0.21 & 0.41 \\
B4\_S3 &                         &                        & 38 & 15.56 & 9.20 & 7.42 & 0.20 & 0.41 \\
\midrule
\textbf{$\pi_{B4}$} & & & 35 & 22.22 & 9.07 & 7.43 & 0.21 & 0.41 \\
\midrule

% ---------------------- D2 ----------------------
D2\_S1 & \multirow{3}{*}{164.94} & \multirow{3}{*}{26.88} & 36 & 20.00 & 7.50 & 7.57 & 0.21 & 0.41 \\
D2\_S2 &                         &                        & 24 & 46.67 & 7.50 & 7.48 & 0.21 & 0.42 \\
D2\_S3 &                         &                        & 34 & 24.44 & 7.60 & 7.47 & 0.22 & 0.43 \\
\midrule
\textbf{$\pi_{D2}$} & & & 31.33 & 30.37 & 7.53 & 7.51 & 0.21 & 0.42 \\
\midrule

% ---------------------- D3 ----------------------
D3\_S1 & \multirow{3}{*}{134.94} & \multirow{3}{*}{3.80} & 37 & 17.78 & 7.90 & 7.87 & 0.21 & 0.42 \\
D3\_S2 &                         &                       & 38 & 15.56 & 7.90 & 7.62 & 0.21 & 0.41 \\
D3\_S3 &                         &                       & 36 & 20.00 & 7.60 & 7.55 & 0.21 & 0.42 \\
\midrule
\textbf{$\pi_{D3}$} & & & 37 & 17.78 & 7.80 & 7.68 & 0.21 & 0.42 \\
\midrule

% ---------------------- D4 ----------------------
D4\_S1 & \multirow{3}{*}{147.79} & \multirow{3}{*}{13.68} & 40 & 11.11 & 8.10 & 7.67 & 0.21 & 0.43 \\
D4\_S2 &                         &                        & 40 & 11.11 & 7.90 & 7.49 & 0.21 & 0.42 \\
D4\_S3 &                         &                        & 40 & 11.11 & 7.90 & 7.44 & 0.22 & 0.43 \\
\midrule
\textbf{$\pi_{D4}$} & & & 40 & 11.11 & 7.97 & 7.53 & 0.21 & 0.43 \\
\midrule

% ---------------------- Global π ----------------------
\textbf{$\boldsymbol{\pi}$}
& \textbf{150.88} 
& \textbf{16.06} 
& \textbf{35.25} 
& \textbf{21.67} 
& \textbf{8.81} 
& \textbf{7.62} 
& \textbf{0.21} 
& \textbf{0.42} \\
\bottomrule

\end{tabularx}
\end{table*}

\section{Discussion}
\label{sec:discussion}

This section interprets the experimental results with emphasis on three core aspects: the USV's automated water-sampling performance, its solar-assisted power architecture, and the platform’s autonomous capabilities under real-world disturbances. The discussion also situates our findings relative to prior USV systems, outlines the practical implications of deploying an autonomy-enabled sampler in natural environments, and identifies technical limitations that should guide future development.

\subsection{Automated water sampling}

The proposed USV demonstrated the capability of collecting up to 72 discrete water samples per mission. During field trials at Achocalla Lagoon, 24 samples were successfully retrieved across distributed waypoints \ref{fig:map}, providing sufficient horizontal resolution to detect clear environmental gradients. For example, average water temperature ranged from approximately 10.10~°C at point A3 to 7.50~°C at point D2, while pH varied between 8.10 at A4 and 7.41 at B4 \ref{tab:table_sampling_data}. At each waypoint, three syringes operated in parallel, enabling local replication. For instance, at point D3, the three syringes measured pH values of 7.87, 7.62, and 7.55 \ref{tab:table_sampling_data}. Differences between sites were consistently larger than within-site replicates, indicating that the observed gradients reflect true environmental heterogeneity rather than sampling noise.

Compared to prior USV implementations, the presented system substantially increases spatial sampling density. Existing platforms typically store only a few discrete samples, for example, two 30~mL syringes in \cite{chang2021}, six syringes in \cite{lim2025}, or four 500~mL containers in \cite{katsouras2024}, which limits spatial representativeness and sensitivity to small-scale variability. In contrast, our system enables the collection of 72 individually isolated 45~mL samples in a single mission, representing a significant advancement in horizontal resolution. Some prior works address vertical heterogeneity by sampling at multiple depths (e.g., \cite{ahmad2025}), underscoring the value of depth control. In our trials, all samples were collected at a fixed depth of 30~cm according to local protocol, which we recognize as a limitation that motivates the integration of adjustable vertical profiling in future iterations.

\subsection{Solar-powered autonomy}

The solar energy system, comprising two photovoltaic panels, a charge controller, and dual LiFePO\textsubscript{4} batteries, was designed to extend the USV's operational endurance by stabilizing power delivery to propulsion, sensing, and onboard computation. Prior work has shown that photovoltaic architectures can significantly increase mission duration in autonomous surface vehicles \cite{aissi2020, chen2021}, reducing dependence on fixed charging infrastructure.

Unlike long-endurance solar platforms, the configuration developed here focuses on energy-efficient operation during short- to medium-range sampling missions, where instantaneous power demands from propulsion, LiDAR, RGB-D perception, and the Jetson Orin Nano dominate consumption. Field results indicate that the USV can operate for approximately one hour under full load, with solar input contributing an additional ten minutes of runtime. Although modest compared to fully solar-sustained systems, this result demonstrates that photovoltaic support can stabilize power availability during peak consumption and improve mission reliability, especially in compact robotic platforms with heterogeneous sensing and real-time autonomy stacks.

These findings support the value of solar-assisted designs even when absolute endurance gains are limited. Future iterations may further increase runtime through improved power distribution, adaptive thrust allocation, lighter sampling modules, or higher-efficiency photovoltaic arrays.

\subsection{Robust autonomy and systems integration}

This work contributes to the field of autonomous environmental robotics by demonstrating how perception, localization, planning, embedded control, and mechanical manipulation can be integrated into a unified sampling platform. A key novelty is the distributed embedded control architecture enabled by micro-ROS, which provides deterministic actuation, module-level fault isolation, and scalable coordination across 24 syringe mechanisms, capabilities rarely explored in aquatic sampling robots.

The platform also demonstrated robust autonomy in an unstructured natural environment, relying on fused GPS-RTK/IMU localization, LiDAR and stereo-vision perception, and behavior-tree mission execution to maintain stable performance under environmental disturbances. Importantly, the results highlight the need to tightly couple mechanical sampling actions with navigation and perception, as successful water acquisition depends on precise localization, obstacle avoidance, and synchronized actuation.

\subsection{Limitations and Future Work}

Despite its demonstrated performance, the current USV platform has several limitations that should be addressed in future iterations.

\textbf{Mechanical aspects.} The average sampling duration per syringe (150.88~s) is relatively long. Filling time could be reduced by using higher-torque motors, thereby eliminating or reducing the need for intermediate power transmission stages. This change would simplify the mechanical design, improve reliability, and increase throughput.

\textbf{Operational aspects.} The system currently performs surface sampling at a fixed depth of 30~cm, making it best suited for lentic water bodies. While appropriate for many water quality protocols, this design limits applicability in stratified environments. Incorporating adjustable vertical profiling would expand the platform’s utility for ecological and hydrological studies.

\textbf{Environmental considerations.} External disturbances, including currents, waves, floating debris, wind, animals, and vegetation, can interfere with navigation and affect sampling accuracy. Enhancing robustness in dynamic water bodies (e.g., rivers, rapids) will require improved control strategies, higher-thrust propulsion, and better environmental modeling.

\textbf{Sensing aspects.} The sampling inlet lacks a filtering mechanism, posing a risk of collecting unwanted particulate matter or organisms, particularly in turbid waters. Introducing a removable fine mesh would reduce clogging and improve sample integrity. Additionally, syringe-based sampling may not preserve Dissolved Oxygen (DO) accurately due to air bubbles or headspace formation. Future versions should incorporate electrochemical or optical DO sensors for reliable in situ measurement.

By addressing these mechanical, operational, environmental, and sensing limitations, future versions of the USV can enhance sampling efficiency, extend applicability to diverse aquatic environments, and improve the robustness of autonomous water quality assessment.

\section{Conclusions}
\label{sec:conclusions}
The development of the autonomous solar-powered USV demonstrated its capability to collect water samples under real field conditions, as shown by trials in Achocalla Lagoon. The system ensured discrete, contamination-free samples with precise volume control, validating its suitability for water quality studies using parameters such as pH, turbidity, and conductivity. Energy autonomy, achieved through solar integration, enabled extended operation times and reduced dependence on external charging, increasing the sustainability of long-duration missions in remote environments. Beyond the technical performance, the platform enables dense spatial sampling at a fraction of the human effort required for traditional campaigns, with potential integration into long-term aquatic monitoring programs in resource-limited regions—thereby contributing to better environmental resolution and informed decision-making for society.

The modular syringe-based architecture, capable of producing 72 samples per cycle, represents a significant advance over manual methodologies and other USVs reported in the literature, enabling high-resolution sampling both spatially and temporally. The integration of autonomous navigation, long-range communication, and a web-based data management platform further enhanced robustness, traceability, and applicability in unstructured environments. Although limitations in sampling speed and the absence of vertical profiling were identified, this work constitutes a step forward in automating environmental monitoring. With improvements in depth control, inlet filtering, and mechanical optimization, the USV can evolve into a more versatile platform and serve as the foundation for scalable networks of autonomous systems for long-term environmental monitoring.

\section*{Acknowledgments}
We appreciate the support of Fabio Díaz Palacios and Magdalena Morales Jimenez, who were preliminarily involved in the project and contributed to defining the foundation of the research gap.

\section*{Author contributions}

Conceptualization, M.M., M.F., G.L., S.L., L.A., C.M., M.H., and E.S.; methodology, M.M., M.F., G.L., S.L., L.A., C.M., and M.H.; software, M.M., M.F., G.L., S.L., L.A., C.M., M.H., and E.S.; validation, M.M., M.F., G.L., L.A., C.M., M.H., and E.S.; investigation, M.M., M.F., G.L., S.L., L.A., C.M., and M.H.; resources, E.S.; data curation, G.L., S.L., C.M., and M.H.; writing—original draft preparation, M.M., M.F., G.L., S.L., L.A., and C.M.; writing—review and editing, E.S.; visualization, M.M., M.F., G.L., and C.M.; supervision, E.S.; project administration, E.S.; funding acquisition, E.S. All authors have read and agreed to the published version of the manuscript.

\section*{Financial disclosure}

The research leading to the results presented here received funding from the Government of Canada through the Canadian Fund for Local Initiatives (CFLI 2024), Project No. CFLI-LIMA-BOL-0001.

\section*{Conflict of interest}

The authors declare no potential conflict of interests.

% ============================================================
% Bibliography
% ============================================================
\bibliographystyle{unsrt}
\bibliography{bibliography}

\end{document}